\documentclass[10pt,twocolumn,letterpaper]{article}

\usepackage{cvpr}              

\usepackage{graphicx}
\usepackage{amsmath}
\usepackage{amssymb}
\usepackage{booktabs}
\usepackage{dblfloatfix}

\usepackage{enumitem}
\usepackage[ruled]{algorithm2e}
\usepackage{color}
\usepackage{soul}
\usepackage[accsupp]{axessibility}
\usepackage[pagebackref,breaklinks,colorlinks]{hyperref}

\usepackage[capitalize]{cleveref}
\crefname{section}{Sec.}{Secs.}
\Crefname{section}{Section}{Sections}
\Crefname{table}{Table}{Tables}
\crefname{table}{Tab.}{Tabs.}

\def\cvprPaperID{9990} 
\def\confName{CVPR}
\def\confYear{2022}

\begin{document}


\title{Robust outlier detection by de-biasing VAE likelihoods}

\author{Kushal Chauhan $^{1}$, Barath Mohan U $^{2}$, Pradeep Shenoy $^{1}$, Manish Gupta $^{1}$, Devarajan Sridharan $^{2}$\thanks{Corresponding author}\\
$^{1}$ Google Research \\
\{\tt kushalchauhan, shenoypradeep, manishgupt\}@google.com \\
$^{2}$ Center for Neuroscience, and Computer Science and Automation, Indian Institute of Science \\
\{\tt barathu, sridhar\}@iisc.ac.in \\
}

\maketitle

\begin{abstract}
Deep networks often make confident, yet, incorrect, predictions when tested with outlier data that is far removed from their training distributions. Likelihoods computed by deep generative models (DGMs) are a candidate metric for outlier detection with unlabeled data. Yet, previous studies have shown that DGM likelihoods are unreliable and can be easily biased by simple transformations to input data. Here, we examine outlier detection with variational autoencoders (VAEs), among the simplest of DGMs. We propose novel analytical and algorithmic approaches to ameliorate key biases with VAE likelihoods. Our bias corrections are sample-specific, computationally inexpensive, and readily computed for various decoder visible distributions. Next, we show that a well-known image pre-processing technique -- contrast stretching -- extends the effectiveness of bias correction to further improve outlier detection. Our approach achieves state-of-the-art accuracies with nine grayscale and natural image datasets, and demonstrates significant advantages -- both with speed and performance -- over four recent, competing approaches. In summary, lightweight remedies suffice to achieve robust outlier detection with VAEs.\footnote{Code is available at \url{https://github.com/google-research/google-research/tree/master/vae_ood}.}
\end{abstract}

\vspace{-3mm}
\section{Introduction}
\label{sec:intro}

\begin{figure}
\centering
\includegraphics[width=0.95\linewidth]{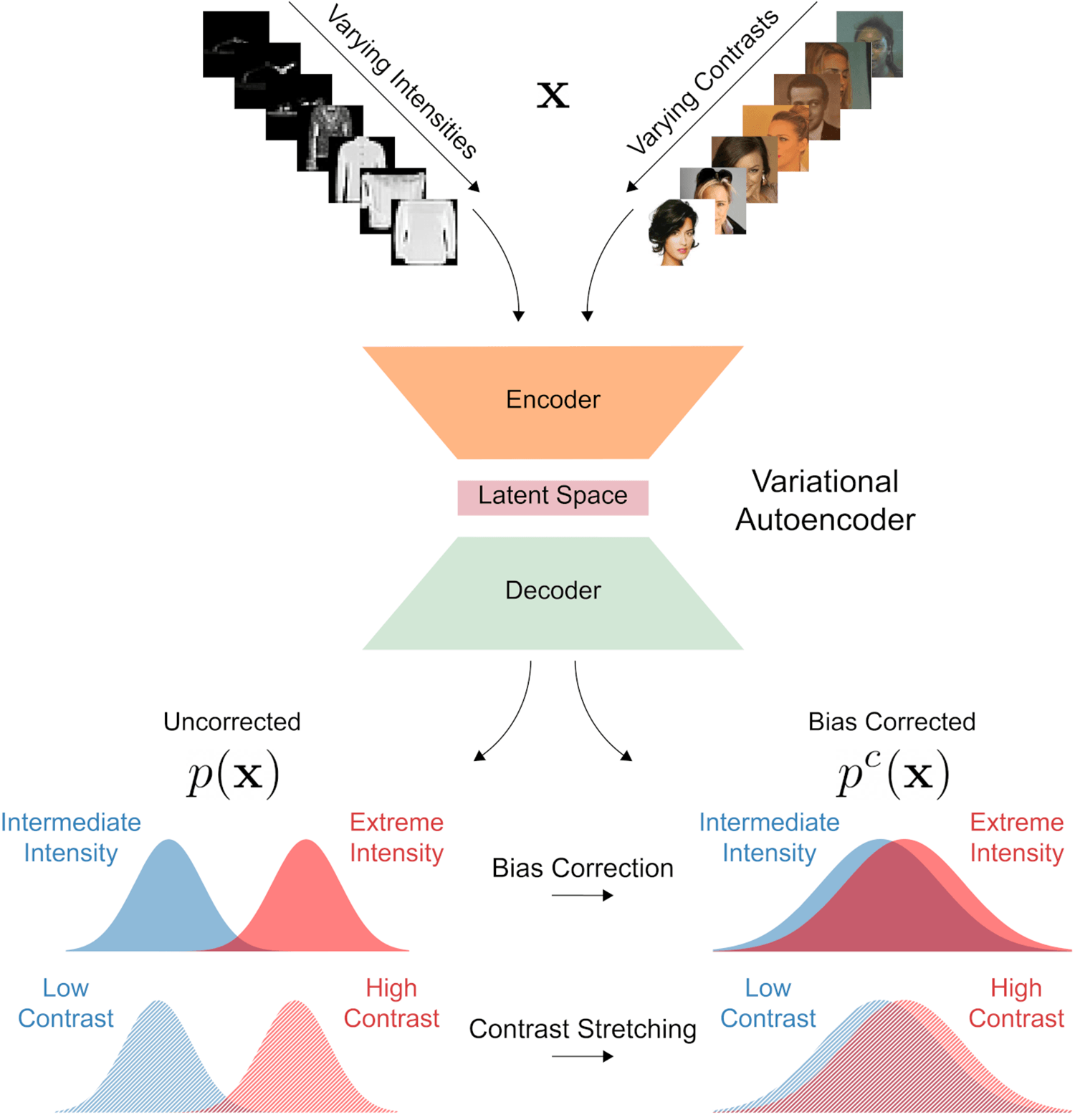}
\caption{
{\bf Schematic of the proposed approach.} Variational autoencoders (VAEs) are routinely employed for unsupervised outlier detection with real-world images. Yet, variations in low-level image features, like overall intensity or contrast, can readily bias VAE likelihoods ($p(\mathbf{x})$). We analyze these biases, and propose  lightweight analytical and algorithmic remedies (Bias Correction and Contrast Stretching) for de-biasing VAE likelihoods ($p^c(\mathbf{x})$), to achieve robust outlier detection.
}
\label{fig:schematic}
\vspace{-1em}
\end{figure}

Deep neural networks are increasingly deployed in real-world computer vision applications. A key issue with such deployments is over-confident predictions: the tendency of these networks to make confident, yet incorrect, predictions when tested with images whose statistics are far removed from the training data distribution~\cite{Snoek2012}. Developing robust methods for outlier detection is, therefore, an important challenge with critical real-world implications.

\begin{figure*}[bhp]
\centering

\begin{subfigure}[b]{0.2\linewidth}
   \includegraphics[width=\textwidth]{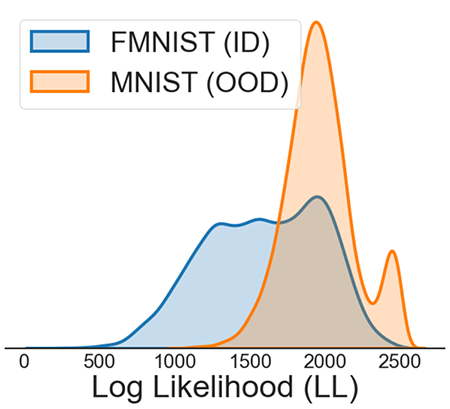}
   \vspace*{-5mm}
   \caption{}
   \label{fig:grayscale_hist}
\end{subfigure}
\hspace{2mm}
\begin{subfigure}[b]{0.15\linewidth}
  \includegraphics[width=\textwidth]{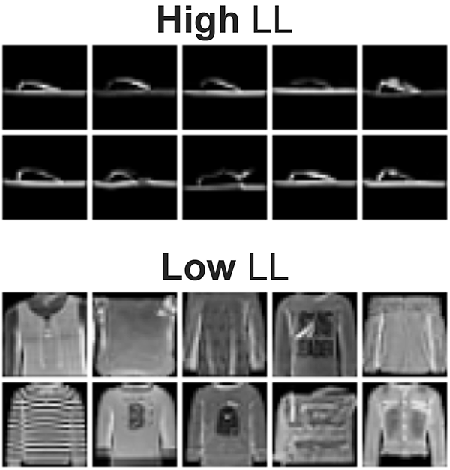}
  \vspace*{-5mm}
  \caption{}
  \label{fig:fmnist_origLL_imgs}
\end{subfigure}
\hspace{2mm}
\begin{subfigure}[b]{0.2\linewidth}
   \includegraphics[width=\textwidth]{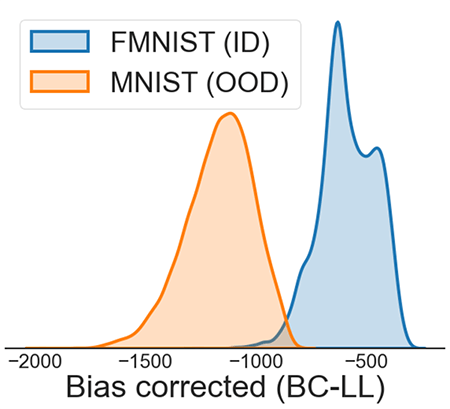}
   \vspace*{-5mm}
   \caption{}
   \label{fig:grayscale_bc_hist}
\end{subfigure}
\hspace{2mm}
\begin{subfigure}[b]{0.15\linewidth}
  \includegraphics[width=\textwidth]{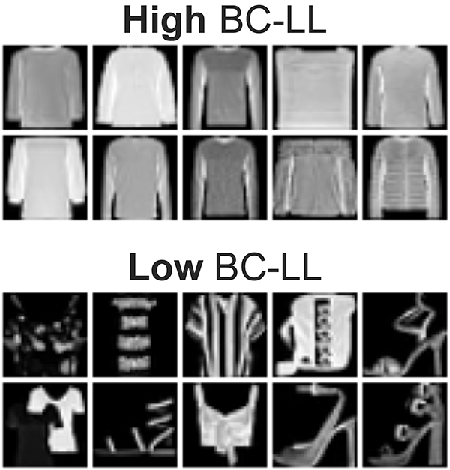}
  \vspace*{-5mm}
  \caption{}
  \label{fig:fmnist_bcLL_imgs}
\end{subfigure}
\hspace{2mm}
\begin{subfigure}[b]{0.2\linewidth}
  \includegraphics[width=\textwidth]{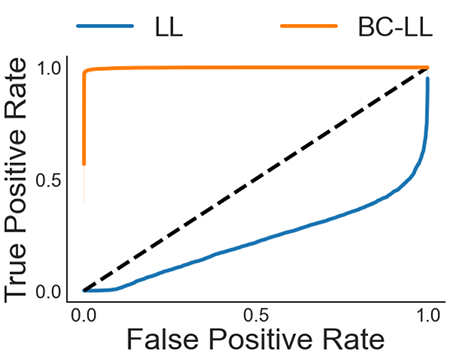}
  \vspace*{-5mm}
  \caption{}
  \label{fig:grayscale_roc}
\end{subfigure}

\begin{subfigure}[b]{0.2\linewidth}
   \includegraphics[width=\textwidth]{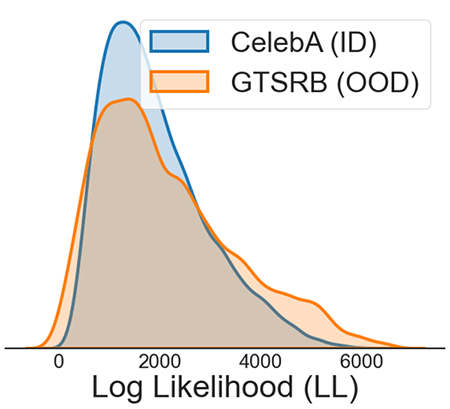}
   \vspace*{-5mm}
   \caption{}
   \label{fig:color_hist}
\end{subfigure}
\hspace{2mm}
\begin{subfigure}[b]{0.15\linewidth}
  \includegraphics[width=\textwidth]{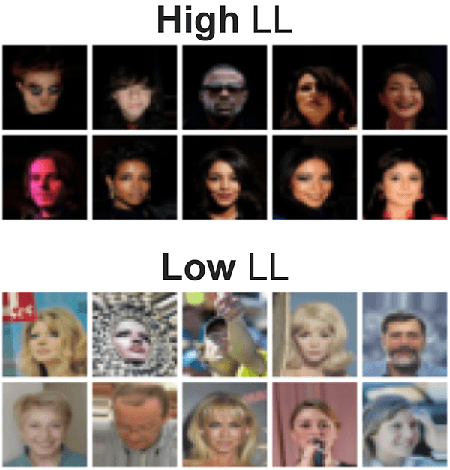}
  \vspace*{-5mm}
  \caption{}
  \label{fig:celeba_origLL_imgs}
\end{subfigure}
\hspace{2mm}
\begin{subfigure}[b]{0.2\linewidth}
   \includegraphics[width=\textwidth]{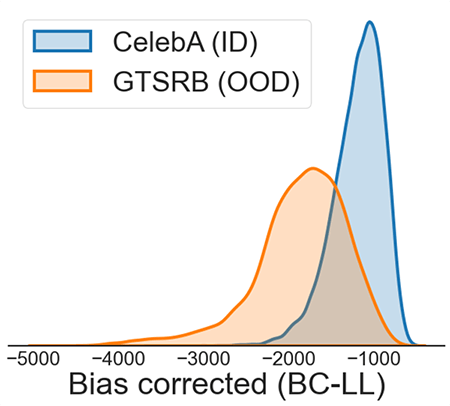}
   \vspace*{-5mm}
   \caption{}
   \label{fig:color_bc_hist}
\end{subfigure}
\hspace{2mm}
\begin{subfigure}[b]{0.15\linewidth}
  \includegraphics[width=\textwidth]{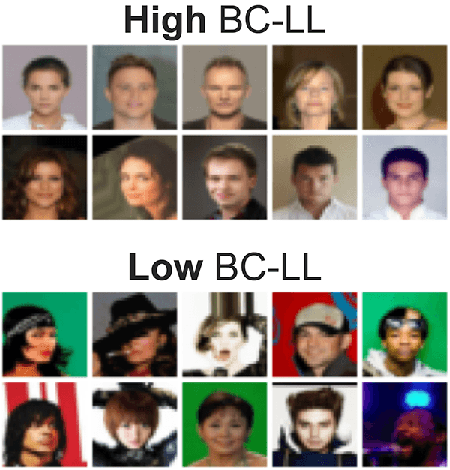}
  \vspace*{-5mm}
  \caption{}
  \label{fig:celeba_bcLL_imgs}
\end{subfigure}
\hspace{2mm}
\begin{subfigure}[b]{0.2\linewidth}
  \includegraphics[width=\textwidth]{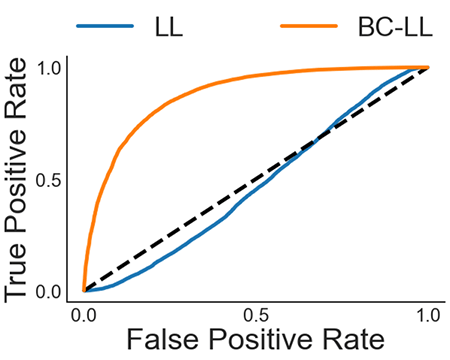}
  \vspace*{-5mm}
  \caption{}
  \label{fig:color_roc}
\end{subfigure}
\caption{{\bf Bias correction improves outlier detection.} {\bf (a)} Kernel density estimate of Fashion-MNIST (FMNIST) VAE log likelihoods. Out-of-distribution (OOD) MNIST samples get higher log likelihoods as compared to in-distribution (ID) FMNIST samples. {\bf (b)} Top and bottom rows: FMNIST images assigned highest and lowest log-likelihoods, respectively, by the FMNIST VAE. {\bf (c)} Same as in panel (a), but with bias correction  {\bf (d)} Same as in panel (b), but with bias correction. {\bf (e)} AUROC curves for outlier detection with the FMNIST VAE likelihoods using MNIST (OOD) test samples without (blue) bias correction and with bias correction (orange)  {\bf (f-j)} Same as in panels (a-e), but for the CelebA VAE with CelebA (ID) and GTSRB (OOD) samples.} 
\label{fig:beforeandafter}
\vspace{-1em}
\end{figure*}

A popular class of approaches for outlier detection, especially with label-free data, involves computing sample likelihoods with deep generative models like variational autoencoders (VAEs~\cite{Kingma2013}) or flow-based models (e.g. Glow~\cite{kingma2018glow}). Here, we explore outlier detection with VAEs, arguably among the simplest of deep generative models. In VAEs, both the generative model $p_{\mathbf{\theta}}(\mathbf{x|z})$ and the inference model $q_{\mathbf{\phi}}(\mathbf{z|x})$ are parameterized by deep neural networks with parameters $\mathbf{\theta}$ (decoder) and $\mathbf{\phi}$ (encoder) respectively; the stochastic latent representation $\mathbf{z}$ is typically of considerably lower dimensionality than the input data $\mathbf{x}$. VAEs differ from other kinds of deep generative models, like flow models, in that rather than directly optimizing the marginal likelihood $p(\mathbf{x})$, they seek to optimize the evidence lower bound (ELBO) as a proxy for maximizing $p(\mathbf{x})$.

Several previous studies have shown that likelihoods computed by deep generative models, including VAEs, are unreliable for outlier detection~\cite{Ren2019,Nalisnick2019,Choi2018,Xiao2020}. For example, these likelihoods are readily biased by differences in low-level image statistics, like the number of zeros in a sample~\cite{Ren2019} or the variance in the distribution of pixel intensities~\cite{Nalisnick2019}. A few solutions have been proposed to overcome these challenges. However, these suffer from computational bottlenecks~\cite{Ren2019, Xiao2020} or do not work well in low-sample regimes~\cite{Nalisnick2019b} (see Section 2, Related Work). 

In this context, we propose and test efficient remedies that achieve or approach state-of-the-art for outlier detection. Our key contributions are as follows:
\vspace{-0.5mm}
\begin{itemize}[leftmargin=*,noitemsep]
    \item We present a careful analysis of bias in VAE likelihoods and propose analytical and algorithmic approaches to correct this bias. These corrections can be computed inexpensively, {\it post hoc}, during evaluation time.
    \item We show that a standard image pre-processing step (contrast normalization) enables competitive outlier detection.
    \item We present a comprehensive evaluation of outlier detection with VAEs, with nine datasets (four grayscale and five natural image datasets), in all. 
    \item We demonstrate key advantages in terms of both speed and accuracy over multiple competing, state-of-the-art approaches~\cite{Serra2019,Ren2019,Xiao2020,Choi2018}. 
\end{itemize}
In sum, lightweight and computationally efficient remedies  suffice to achieve robust outlier detection with VAEs.

\section{The challenge of outlier detection with VAE likelihoods}

We briefly recapitulate a well-known challenge of outlier detection with VAE likelihoods, by taking a fresh look at two previously reported sources of bias~\cite{Ren2019,Nalisnick2019}.

{\it Bias arising from pixel intensity}. As a first example, we train a VAE on grayscale FMNIST images ~\cite{FashionMNIST2017} and compute the likelihoods for in-distribution (ID) FMNIST and out-of-distribution (OOD) MNIST~\cite{MNIST2010} test samples. We employ the continuous Bernoulli visible distribution \cite{Loaiza-Ganem2019} for the VAE decoder (model details, Appendix A. We replicate the well-known issue with VAE likelihoods: FMNIST VAE likelihoods are higher for OOD (MNIST) samples as compared to ID (FMNIST) samples (Fig.~\ref{fig:grayscale_hist}). The highest likelihoods are assigned to FMNIST samples with a large number of black pixels (Fig.~\ref{fig:fmnist_origLL_imgs}, top row), whereas the lowest likelihoods are assigned to samples with many intermediate (gray) pixel values (Fig.~\ref{fig:fmnist_origLL_imgs}, bottom row), consistent with previous reports~\cite{Ren2019}. On simulated images with different (constant) pixel intensities, we find a U-shape trend in likelihood bias (Fig.~\ref{fig:intensity_bias}, gray line).

{\it Bias arising from channel variance or image contrast}. Next, we train a VAE on the CelebA dataset~\cite{CelebA2015}, and compute ID (CelebA) and OOD  (GTSRB~\cite{GTSRB2012}) likelihoods. Again, the VAE assigns high likelihoods to OOD samples  (Fig.~\ref{fig:color_hist}). Faces with dark backgrounds and high contrast between the face and background are assigned the highest likelihoods (Fig.~\ref{fig:celeba_origLL_imgs}, top row), and vice versa for low contrast faces (Fig.~\ref{fig:celeba_origLL_imgs}, bottom row). With simulated images, we observe that VAE likelihoods are strongly biased by contrast (Fig.~\ref{fig:contrast_bias}, gray line).

In this context, we develop efficient remedies for outlier detection with VAE likelihoods. First, we correct for pixel intensity bias using both analytical and empirical approaches. Second, we correct for the image contrast bias using a standard image pre-processing step (contrast stretching). Finally, we evaluate our remedies with multiple (nine) grayscale and natural image datasets, to show state-of-the-art outlier detection.

\subsection*{Related work}
Our bias correction is most closely related to the work of Serra et al.\ (2019)~\cite{Serra2019} who proposed a correction for Glow and PixelCNN++ model likelihoods based on ``input complexity'' (IC). Their out-of-distribution score is computed by subtracting a sample-specific complexity estimate $L(\mathbf{x})$ from the negative log-likelihood. Yet, IC depends on an {\it ad hoc} choice of a compression algorithm (e.g. PNG, JPEG2000, FLIF), whereas our correction is analytically derived from the VAE decoder visible distribution. We also demonstrate superior performance over IC for multiple grayscale and natural image datasets.

Ren et al.\ (2019)~\cite{Ren2019} originally highlighted the problem of bias in deep generative model likelihoods, for samples with many zero-valued pixels. They proposed correcting for this bias by training a second generative model with noise-corrupted samples to capture background statistics; the ``likelihood ratio'' between the original and noisy VAEs provided a sensitive readout of foreground object statistics. Yet, this approach does not appear to work well with VAEs (see also~\cite{Xiao2020}). Moreover, our bias correction obviates the need for training multiple, duplicate models. Similarly, Nalisnick et al.\ (2019)~\cite{Nalisnick2019} originally identified the problem of bias in likelihoods arising from sample variance~\cite{Nalisnick2019b}. However, their solution -- a ``typicality'' test -- works best with batches of samples, and performs relatively poorly with single outlier samples. 


Recent work by Yong et al.\ (2020)~\cite{Yong2020} proposed employing bias-free Gaussian likelihoods and their variances for outlier detection. Yet, Gaussian visible distributions are not theoretically appropriate to model the finite range of pixel values (0-255) encountered in images. We develop bias corrections for Bernoulli, continuous Bernoulli, truncated Gaussian and categorical visible distributions, which are appropriate for modeling pixel values in images.
Moreover, Yong et al.'s approach does not work well with natural image datasets (their Appendix C). Our results show robust outlier detection, even with natural image datasets.

Xiao et al.\ (2020)~\cite{Xiao2020} proposed a ``likelihood regret'' metric that involves quantifying the improvement in marginal likelihood by retraining the encoder network to obtain optimal likelihood for each sample. Such sample-specific optimizations are computationally expensive, for example, when millions of samples need to be evaluated on-the-fly. In contrast, our proposed metrics are readily computed with a single forward pass through pre-trained VAEs, leading to a ~50-100x speedup in evaluation times over likelihood regret.
\section{De-biasing VAE likelihoods}

To improve outlier detection with VAE likelihoods, we develop remedies for correcting for the two sources of bias discussed in the previous section.

\subsection{Analytical correction for intensity bias}

We develop an analytically-derived correction for VAE likelihoods, based on the recently developed continuous Bernoulli visible distribution for the decoder~\cite{Loaiza-Ganem2019}. The procedure for computing the correction for a conventional Bernoulli visible distribution is presented in Appendix B.

The VAE marginal likelihood can be written as:
$$
\log p_{\mathbf{\theta}}(\mathbf{x}) = \log \mathbb{E}_{q_{\mathbf{\phi}}(\mathbf{z|x})} [ p_{\mathbf{\theta}}(\mathbf{x|z}) p(\mathbf{z}) /  q_{\mathbf{\phi}}(\mathbf{z|x}) ]
$$

\begin{figure}[tp]
\centering
\begin{subfigure}[b]{0.49\linewidth}
   \includegraphics[width=\textwidth]{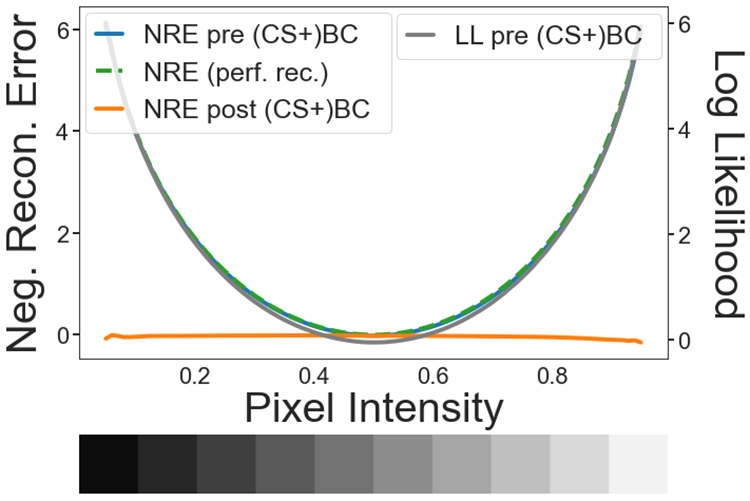}
   \caption{}
    \label{fig:intensity_bias}
   
\end{subfigure}
\hspace{0mm}
\begin{subfigure}[b]{0.49\linewidth}
   \includegraphics[width=\textwidth]{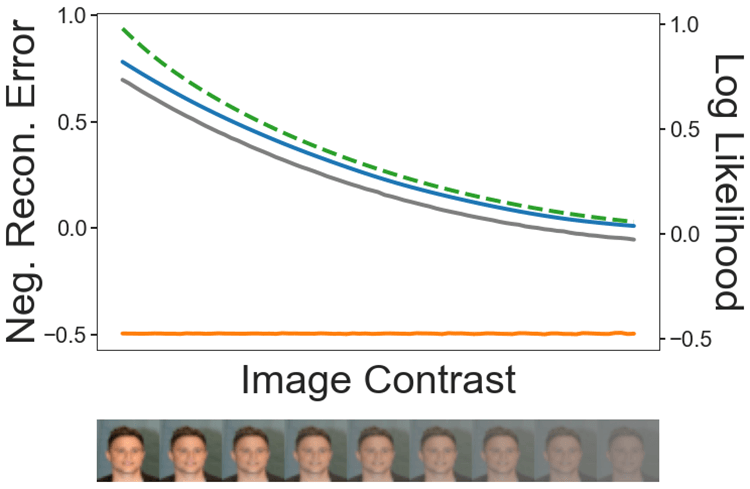}
   \caption{}
 \label{fig:contrast_bias}

\end{subfigure}
\begin{subfigure}[b]{0.49\linewidth}
   \includegraphics[width=\textwidth]{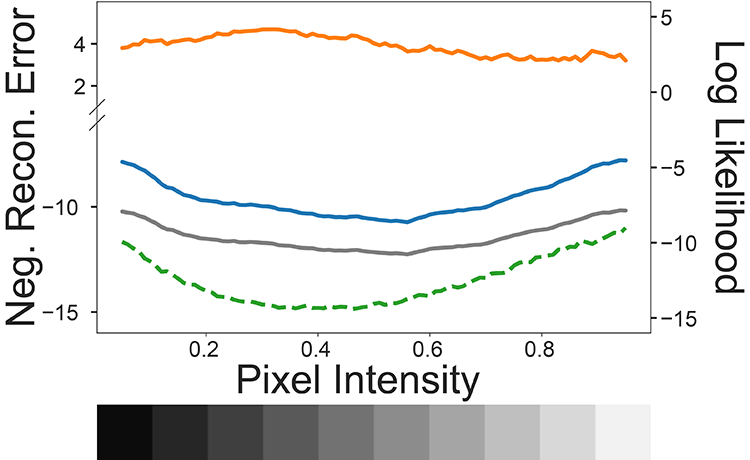}
   \caption{}
\label{fig:intensity_bias_cat}

\end{subfigure}
\hspace{0mm}
\begin{subfigure}[b]{0.49\linewidth}
   \includegraphics[width=\textwidth]{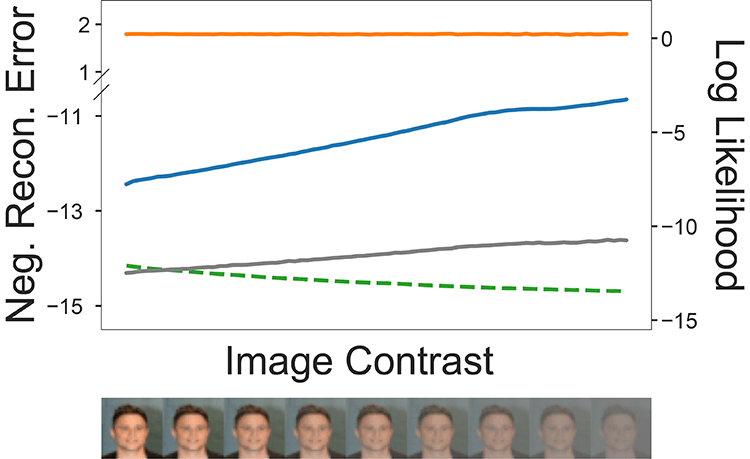}
   \caption{}
\label{fig:contrast_bias_cat}

\end{subfigure}
\caption{\textbf{Correcting for biases arising from pixel intensities and image contrasts.} {\bf (a)} Negative reconstruction error for simulated images with uniform pixel intensities, ranging from full-black to full-white before (blue) and after (orange) analytical bias correction (BC) (y-axes units are kilonats). {\bf (b)} Negative reconstruction error for simulated images with varying contrasts, from highest to lowest values before (blue) and after (orange) contrast stretching (CS) and analytical bias correction. (Both panels) Dashed green line: negative reconstruction error for perfect reconstruction with the continuous Bernoulli visible distribution. Gray line: log-likelihood before bias correction. {\bf (c-d)} Same as in (a) and (b) but for VAE trained with categorical visible distribution, and algorithmic bias correction.}
\label{fig:bias}
\vspace{-1em}
\end{figure}

We examine the negative reconstruction error term $p_{\mathbf{\theta}}(\mathbf{x|z})$, assuming perfect reconstruction of the input samples by the VAE. We denote this as $ p_{c\textsc{b}}(\mathbf{x; \boldsymbol{\lambda}^*})$ where $p_{c\textsc{b}}$ denotes the continuous Bernoulli pdf, and $\boldsymbol{\lambda}^*$ are optimal parameters that correspond to perfect reconstruction ($\mathbf{\hat{x}} = \mathbf{x}$). We plot $\log p_{c\textsc{b}}(\mathbf{x; \boldsymbol{\lambda}^*})$ for simulated images in Figure~\ref{fig:intensity_bias} (dashed green). $\log p_{c\textsc{b}}(\mathbf{x; \boldsymbol{\lambda}^*})$ exhibits a bias that is nearly identical with the marginal likelihood (Fig.~\ref{fig:intensity_bias}, gray). Thus, even if two input samples are perfectly reconstructed by the VAE, these will be assigned different likelihoods, depending on the average pixel intensity in each sample; a bias that is largely driven by the reconstruction error term.

We eliminate this bias in the reconstruction error by dividing by the error for perfect reconstruction. For the continuous Bernoulli visible distribution, the negative reconstruction error is given by:
\begin{align} \label{eq_llcontbern}
\begin{split}
\log p_\theta (\mathbf{x} | \mathbf{z}) \ \ \ \ \ &= \ \ \ \  \log p_{c\textsc{b}}(\mathbf{x}; \boldsymbol{\lambda}_\theta(\mathbf{z})) \\ 
= \sum_{i=1}^D \log C(\lambda_i) & + x_i \log \lambda_i + 
 (1-x_i) \log (1-\lambda_i)
\end{split}
\end{align}

Note that the continuous Bernoulli decoder outputs the shape parameter ($\lambda_i$) for the $i^\mathrm{th}$ pixel. The decoded pixel value itself is given by: 
\begin{align*}
\hat{x}_i &= \frac{\lambda_i}{2\lambda_i -1} + \frac {1}{2\tanh^{-1}(1-2\lambda_i )} & \mathrm{ if } \ \ \ \ \lambda_i \neq \frac {1}{2} \\
 \ &= \frac {1}{2} & \mathrm{ if } \ \ \ \ \lambda_i = \frac {1}{2}
\end{align*}

For perfect reconstruction we set  $\hat{x}_i=x_i$. To find the optimal $\lambda^*_i$ corresponding to perfect reconstruction, we used SciPy's implementation of Nelder-Mead simplex algorithm to iteratively maximize $\log p_{c\textsc{b}}(x_i; \lambda_i)$; the correction is then calculated by setting $\lambda_i = \lambda^*_i$ in equation (\ref{eq_llcontbern}) above.

Thus, the ``bias-corrected'' marginal likelihood (BC) evaluates to: 
\begin{align} \label{eq_biascorrLL}
\log p^c_{\mathbf{\theta}}(\mathbf{x}) &=
\log \mathbb{E}_{q_{\mathbf{\phi}}(\mathbf{z|x})}  \left[ \frac{p_{\mathbf{\theta}}(\mathbf{x|z})}{p_{c\textsc{b}}(\mathbf{x; \boldsymbol{\lambda}^*})} 
\frac{p(\mathbf{z})} {q_{\mathbf{\phi}}(\mathbf{z|x})} \right] \\
 &= \log p_{\mathbf{\theta}}(\mathbf{x}) - \log p_{c\textsc{b}}(\mathbf{x; \boldsymbol{\lambda}^*})  
\end{align}

Following this analytical bias correction,  the bias in the negative reconstruction error is eliminated (Fig.~\ref{fig:intensity_bias}, orange). We note that this correction can be computed during evaluation time and does not require retraining the VAE. 

\subsection{Algorithmic correction for intensity bias}
Next, we demonstrate an algorithmic approach for correcting biases with VAE likelihoods. Such an approach is relevant for decoder visible distributions for which the analytical correction is not tractable, or for which biases must be evaluated empirically. We illustrate this latter scenario with the ``categorical'' visible distribution'' -- a popular choice of visible distribution for various generative models, including VAEs~\cite{Xiao2020, Ren2019} and autoregressive models like PixelCNN~\cite{pixelcnn}. 

In theory, the categorical distribution does not suffer from an analytical bias. As a result, perfectly reconstructed images should not suffer from any bias in the reconstruction error term, $p_\theta (\mathbf{x} | \mathbf{z})$.
Yet, in practice, we find such a bias arises in a VAE trained with the categorical visible distribution for the decoder. 

To illustrate this, we compute the empirical bias in the log likelihood for a CelebA VAE for images with various uniform pixel intensities (0-255) (Fig.~\ref{fig:intensity_bias_cat}, gray). We observe a U-shaped profile, similar to that observed for VAEs trained with the continuous Bernoulli visible distribution (Fig.~\ref{fig:intensity_bias}). This bias can be explained as follows: For target pixel values close to full black or full white, the VAE decoder concentrates probability mass (PMF) at the target pixel value. On the other hand, for target values close to the middle of the gray range, the PMF is more dispersed around the target pixel value, resulting in lower PMF at the target pixel itself (Appendix F.1, for details). 

We correct for this empirical bias with the following algorithmic approach. We compute the average categorical distribution output for every target pixel value (0-255) employing all of the training (inlier) samples for that VAE. Then, for each test sample, we compute the correction term for the negative reconstruction error as its average value across pixels, under the categorical visible distribution. The procedure is described in detail in Algorithm~\ref{algo:corr_cat}.

\begin{algorithm}[h!]
\DontPrintSemicolon
\SetKwFunction{Mean}{Mean}
\SetKwFunction{Log}{Log}
\KwData{Training Set $\mathbf{X} = \{\mathbf{x_1}, \mathbf{x_2}, ... \mathbf{x_n}\}$ with $\mathbf{x_p}$ of shape $32 \times 32 \times nc$ (no. of channels), Encoder Parameters $\phi$, and
Decoder Parameters $\theta$}
\KwResult{Log Correction Factor $\mathbf{C}: (v, k) \rightarrow$ \textit{Float} for $v = 0,1, \ldots 255$ and $k = 1,2, \ldots nc$}
\SetKwInput{KwInit}{Init}
\BlankLine
\KwInit{Map $\mathbf{A}: (v, k) \rightarrow$ \textit{EmptyList} for $v = 0,1, \ldots 255$ and $k = 1,2, \ldots nc$}

\For{$\mathbf{x_p} \in \mathbf{X}$}{
    \KwInit{Map $\mathbf{B}: (v, k) \rightarrow$ \textit{EmptyList} for $v = 0,1, \ldots 255$ and $k = 1,2, \ldots nc$}
    $\mathbf{z} \sim q_{\phi}(\mathbf{z}|\mathbf{x_p})$
    
    \For{$i \leftarrow 1$ \KwTo $32$, $j \leftarrow 1$ \KwTo $32$, $k \leftarrow 1$ \KwTo $nc$}{
    Append $p_{\theta}^{ijk}(x_p^{ijk} | \mathbf{\bar{z}})$ to $\mathbf{B}(x_p^{ijk}, k)$
    }
    
    \For{$v \leftarrow 0$ \KwTo $255$, $k \leftarrow 1$ \KwTo $nc$}{
     Append $\Mean(\mathbf{B}(v, k))$ to $\mathbf{A}(v, k)$\;
    }
}
\BlankLine
\KwInit{Map $\mathbf{C}: (v, k) \rightarrow 0$ for $v = 0,1, \ldots 255$ and $k = 1,2, \ldots nc$}
\For{$v \leftarrow 0$ \KwTo $255$, $k \leftarrow 1$ \KwTo $nc$}{
    $\mathbf{C}(v, k) \leftarrow \Log(\Mean(\mathbf{A}(v, k)))$
}
\caption{Algorithmic Bias correction}
\label{algo:corr_cat}
\end{algorithm}

Again, following this algorithmic bias correction, the empirical bias with the categorical distribution is eliminated (Fig.~\ref{fig:intensity_bias_cat}, orange). Such an algorithmic correction generalizes to other types of visible distributions as well: results for VAEs trained with the truncated Gaussian visible distribution are presented in Appendix F.2. 

\subsection{Correction for contrast bias: Normalization} \label{sec:cs}
Variation in image contrasts produces systematic biases in the likelihood and negative reconstruction error, both with the continuous Bernoulli visible distribution (Fig.~\ref{fig:contrast_bias}, blue), as well as with the categorical visible distribution (Fig.~\ref{fig:contrast_bias_cat}, blue). This is particularly surprising because the categorical distribution is, in theory, bias-free. Nonetheless this bias arises, in practice, due to practical limitations with achieving perfect reconstructions with VAEs. Specifically, the VAE achieves systematically worse reconstructions -- as measured by the categorical negative reconstruction error -- for  high-contrast, as compared to low-contrast, images (Fig.~\ref{fig:contrast_bias_cat}, blue line). As a consequence, the contrast-related bias in the categorical distribution is in a direction opposite to that observed with the continuous Bernoulli distribution (compare Fig.~\ref{fig:contrast_bias_cat} with Fig.~\ref{fig:contrast_bias}). 

For eliminating this additional source of bias arising from  image contrasts, we propose a standard image pre-processing step -- ``contrast stretching'' -- which is known to improve the robustness of deep classifier models~\cite{Hendrycks2019}. For our case, each image sample, both from the training and testing datasets, is contrast normalized with the following  transformation: $
x_i = \min(\max (0, [x_i - a]/r), 1) $, where $x_i$ refers to the $i^\mathrm{th}$ pixel of image $\mathbf{x}$,
$r = P_{95}(\mathbf{x}) - P_{5}(\mathbf{x})$, $a = P_{5}(\mathbf{x})$, $P_j$ refers to the $j^\mathrm{th}$ percentile and $\mathbf{x}$ refers to the vectorized input sample tensor. Following contrast stretching, we observe more homogeneous distributions of per-channel variance across datasets (see Appendix E.4). 

Contrast stretching and bias correction  ameliorate this bias for both types of visible distributions (Fig.~\ref{fig:contrast_bias}, orange, Fig.~\ref{fig:contrast_bias_cat}, orange).

\section{Experiments}

\begin{figure*}
\centering
\includegraphics[width=0.95\textwidth]{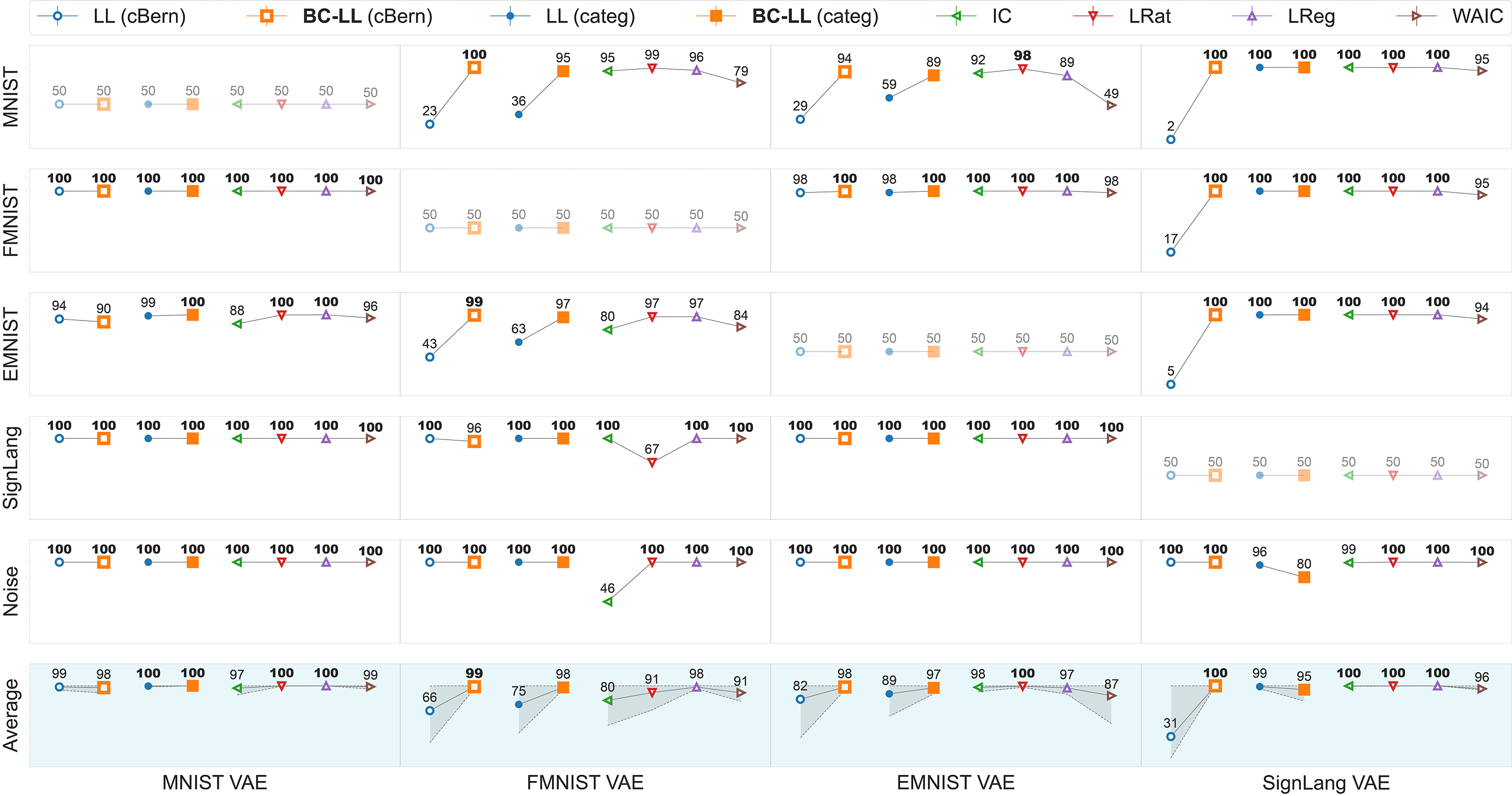}
\caption{\textbf{Outlier detection with bias-corrected likelihoods, and comparison with competing approaches: Grayscale datasets.} Outlier detection AUROC values for VAEs trained with grayscale image datasets and tested with other grayscale datasets (or noise). Each column represents a training dataset, and each row represents a test dataset. Last row: average AUROC across all test datasets for each VAE. Unfilled blue and orange symbols: AUROC with uncorrected log likelihoods (LL), and bias-corrected log likelihoods (BC-LL) respectively for VAE trained with continuous Bernoulli visible distribution. Filled blue and orange symbols: Same as the unfilled symbols but for VAE trained categorical visible distribution. Green, red, purple, and brown symbols show AUROC scores of competing outlier detection approaches: Input Complexity (IC), Likelihood Ratio (LRat), Likelihood Regret (LReg), and Watanabe Akaike Information Criterion (WAIC), respectively. Higher values indicate better outlier detection performance. Numbers in bold indicate highest AUROC values across all scores for each train-test combination. Gray shading in the last row: range of AUROC values for each score across all test datasets.}
\label{fig:grayscale}
\vspace{-1em}
\end{figure*}

We trained and tested VAEs with multiple grayscale and natural image datasets. These included four grayscale image datasets -- MNIST, Fashion-MNIST, EMNIST (handwritten English letters~\cite{EMNIST2017}) and Sign Language-MNIST (hand gesture thumbnails~\cite{hsmnist}) -- and tested them against four outlier datasets each. These outlier datasets comprised each of the other three datasets (the ones that the VAE was not trained on) and a grayscale, uniform noise dataset. Similarly, we trained VAEs with five natural image datasets -- SVHN (Street View House Numbers~\cite{SVHN2011}), CelebA, ComprehensiveCars (traffic surveillance camera views of cars~\cite{CompCars2015}), GTSRB (German Traffic Sign Recognition Benchmark) and CIFAR-10~\cite{CIFAR102009} -- and tested them against five outlier datasets each. Again, the outlier datasets comprised each of the other four datasets, and a colored, uniform noise dataset. Data sources and pre-processing details are provided in Appendix C. Each VAE was trained 6 times, with 3 different random initializations of the network weights across 2 train-validation splits; we report  average performance measures across all 6 runs (e.g. Figs.~\ref{fig:grayscale}, \ref{fig:natural}).

\subsection{De-biased likelihoods improve outlier detection}

We compared the outlier detection performance of the vanilla (uncorrected) VAE likelihood (LL) against the bias corrected likelihood (BC-LL). Both train and test images were contrast stretched prior to computing BC-LL scores. The results are shown as a grid (4$\times$5 or 5$\times$6) of area under the ROC curve (AUROC) values, separately for the grayscale (1 channel) and natural image (3 channel) datasets. Corresponding results for area under the precision-recall curve (AUPRC) and false-positive rate at 80\% true-positive rate (FPR@80\%TPR) are provided in Appendices E.2-E.3, and details regarding each measure are provided in Appendix D.

We evaluated the effectiveness of the analytical bias correction, first, with VAEs trained with the continuous Bernoulli visible distribution.  

First, we evaluated our remedies for outlier detection with VAEs trained on grayscale image datasets. In nearly every case, we found that bias-corrected likelihoods (BC-LL, Fig.~\ref{fig:grayscale}, unfilled orange squares) outperformed  uncorrected likelihoods (LL, Fig.~\ref{fig:grayscale}, unfilled blue circles); AUROC values for outlier detection approached ceiling levels, in many cases. For example, bias correction resolved the problematic case of the Fashion-MNIST VAE such that in-distribution Fashion-MNIST samples were assigned higher bias-corrected likelihoods than out-of-distribution MNIST samples (Fig.~\ref{fig:grayscale_bc_hist}), resulting in a perfect AUROC (Fig.~\ref{fig:grayscale_roc}). Moreover, samples that were assigned the highest and lowest bias-corrected likelihoods  were visually more typical and atypical, respectively (Fig.~\ref{fig:fmnist_bcLL_imgs}), as compared to those based on uncorrected likelihoods (Fig.~\ref{fig:fmnist_origLL_imgs}). Across all outlier datasets tested with the Fashion-MNIST VAE, AUROC was typically at or near ceiling with the bias-corrected likelihoods (Fig.~\ref{fig:grayscale}, second column). On average, we found accuracy improvements of up to $\mathtt{\sim}$70\% following bias correction, across all grayscale VAEs (Fig.~\ref{fig:grayscale}, last row). 

\begin{figure*}[bth]
\centering
\includegraphics[width=0.95\textwidth]{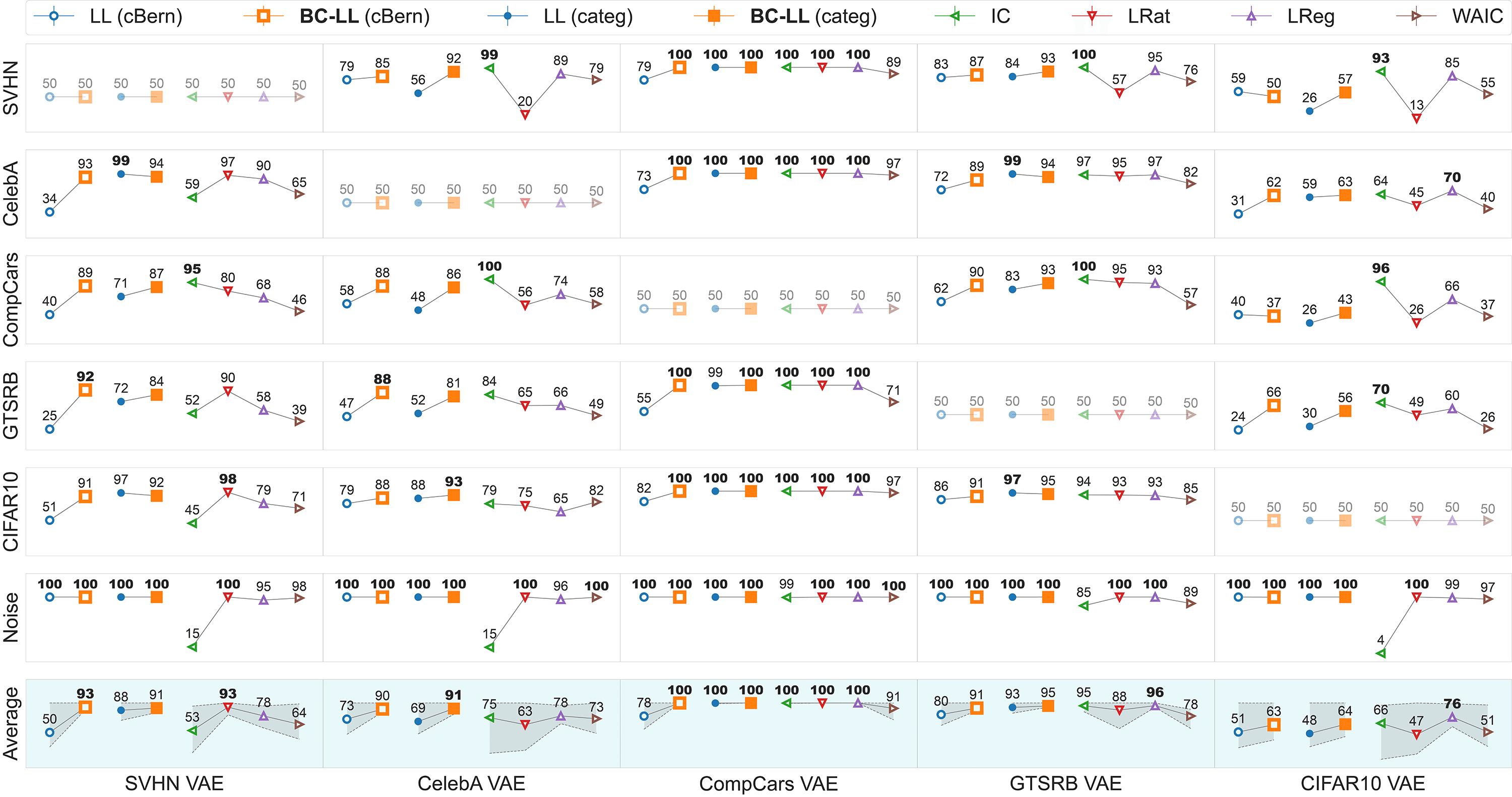}
\caption{
{\bf Outlier detection with bias-corrected likelihoods, and comparison with competing approaches: Natural image datasets.}
Same as in Figure~\ref{fig:grayscale} but for VAEs trained with natural image datasets and tested with other natural image datasets (or noise). Other conventions are the same in Figure~\ref{fig:grayscale}.}
\label{fig:natural}
\vspace{-1em}
\end{figure*}

Second, we tested these remedies with VAEs trained on natural image datasets. Again, in nearly every case the BC-LL score (Fig.~\ref{fig:natural}, unfilled orange squares) outperformed  uncorrected likelihoods (Fig.~\ref{fig:natural}, unfilled blue circles). For example, for the CelebA VAE, AUROC values ranged between 47-79 with the uncorrected likelihoods (across all out-of-distribution datasets, except Noise), whereas these values improved to 85-88 with bias correction  (Fig.~\ref{fig:natural}, second column,  Fig.~\ref{fig:color_bc_hist}, and Fig.~\ref{fig:color_roc}). As with the FashionMNIST dataset, CelebA samples assigned the highest bias-corrected likelihoods were visually more typical; containing face images on typical plain backgrounds, whereas the lowest likelihood samples comprised of face images with extraneous objects, such headgear, unconventional attire or unusual backgrounds (Fig.~\ref{fig:celeba_bcLL_imgs}). On average, we observed accuracy improvements of $\mathtt{\sim}$10-40\% following bias correction, across all natural image VAEs (Fig.~\ref{fig:natural}, last row). 

\vspace{-0.1mm}
Ablation experiments revealed the relative contributions of contrast stretching and bias correction; the results are presented in Appendix E.4. First, contrast stretching was essential to achieving high accuracies, particularly with the natural image datasets. Moreover, contrast stretching, even at test time, sufficed to achieve robust outlier detection: With a VAE trained on the original images (without contrast stretching), outlier detection accuracies improved if contrast stretching was applied at test time alone. Lastly, contrast normalization with techniques other than contrast stretching (e.g. histogram equalization) also yielded comparable outlier detection accuracies (see Appendix E.4). 

\vspace{-0.05mm}
Contrast stretching and bias correction generally improved outlier detection even with milder perturbations to the data, such as with adding various kinds of noise (Gaussian, impulse etc; see Appendix E.5). On the other hand, training the VAE by introducing intensity and contrast variations in the training samples was not as effective as contrast stretching and bias correction (Appendix E.6). Similarly, an alternative approach for normalizing per channel variance using image whitening (Zero-phase Component Analysis or ZCA) was also comparatively unsuccessful, especially with natural image datasets (Appendix E.7).

\vspace{-0.05mm}
Finally, we replicated these results with the algorithmic bias correction with VAEs trained with the categorical visible distribution. Again, in nearly every case we observed significant improvements in outlier detection performance following bias correction (Fig.~\ref{fig:grayscale} and Fig.~\ref{fig:natural}, filled blue circles/LL versus filled orange squares/BC-LL).

\subsection{Comparison with competing approaches}

We compared accuracies with the bias-corrected likelihood (BC-LL) scores against four competing, state-of-the-art approaches: i) input complexity (IC), ii) likelihood ratio (LRat), iii) likelihood regret (LReg) and iv) WAIC (Watanabe-Akaike Information Criterion). Briefly, IC employs a subtractive bias correction for each sample; the correction is computed based on sample complexity estimated with one of many compression algorithms (e.g. PNG)~\cite{Serra2019}. The likelihood ratio score computes the ratio between the sample likelihood obtained with a VAE trained with the original data, and one trained with noise-corrupted images~\cite{Ren2019}. Likelihood regret computes the improvement in the log-likelihood for a particular sample that can be achieved with a sample-specific optimization of the posterior $q_\phi(\mathbf{z}|\mathbf{x})$~\cite{Xiao2020}. Finally, WAIC is computed as $\mathbb{E}_\theta [\log p_\theta (\mathbf{x}) ] - \mathrm{Var}_\theta [\log p_\theta (\mathbf{x}) ]$, from an ensemble of deep generative models~\cite{Choi2018} (Appendix D, for details). 

First, we performed an exhaustive (4$\times$5) comparison of approaches with grayscale VAEs and datasets. For illustration, we discuss results for the problematic Fashion-MNIST VAE (Fig.~\ref{fig:grayscale}, second column).
The BC-LL score outperformed, or performed comparably with, other state-of-the-art methods for detecting outliers with the Fashion-MNIST VAE. Interestingly, IC performed relatively poorly for this VAE (Fig.~\ref{fig:grayscale}, green symbols). Similar results were observed with the other grayscale VAEs: on average, BC-LL AUROC values were comparable with these state-of-the-art approaches (Fig.~\ref{fig:grayscale}, bottom row, Average). 

Next, we performed an exhaustive (5$\times$6) comparison of all approaches with natural image VAEs and datasets. Again, we illustrate the results with the CelebA VAE (Fig.~\ref{fig:natural}, second column).
As before, we noticed superlative outlier detection performance with the BC-LL score. Although, other methods (e.g. IC or LReg) performed better at detecting outliers belonging to specific datasets (e.g. SVHN, Fig.~\ref{fig:natural}, top row, second column),  BC-LL yielded the highest average AUROC values across all outlier datasets tested with the CelebA VAE (Fig.~\ref{fig:natural}, bottom row, second column). Overall, across all combinations of VAEs and outlier datasets tested, BC-LL performed more consistently and yielded AUROC values on par with, or exceeding, state-of-the-art (Fig.~\ref{fig:natural}, bottom row, Average). 

Interestingly, outlier detection with BC-LL was relatively poor for the CIFAR-10 dataset, a failure shared by other approaches as well. Even IC that achieved high AUROC values for the challenging case of the CIFAR-10/ID versus SVHN/OOD case, performed sub-par when tested with the opposite pairing (SVHN/ID vs CIFAR-10/OOD) (Fig.~\ref{fig:natural}, compare first row, last column with fifth row, first column). Moreover, IC's performance significantly deteriorated when tested against Noise OOD images for multiple ID datasets (e.g. SVHN, CelebA, CIFAR-10; Fig.~\ref{fig:natural}, penultimate row, first, third and fifth columns). Similarly, the likelihood ratio metric that achieved superlative AUROCs for the SVHN/ID versus CelebA/OOD case failed when tested with the opposite pairing (CelebA/ID vs SVHN/OOD). The results indicate an important caveat with claiming superlative outlier detection by evaluating specific, challenging ID/OOD pairs (e.g. CIFAR-10/ID versus SVHN/OOD). It is possible that high outlier detection accuracies may occur for such specific pairings due to ``over-correction'' of the OOD likelihood by the respective methods. We discuss this case in the Limitations section and, in further detail, in Appendix G.

\begin{figure}
\centering
\includegraphics[width=0.95\linewidth]{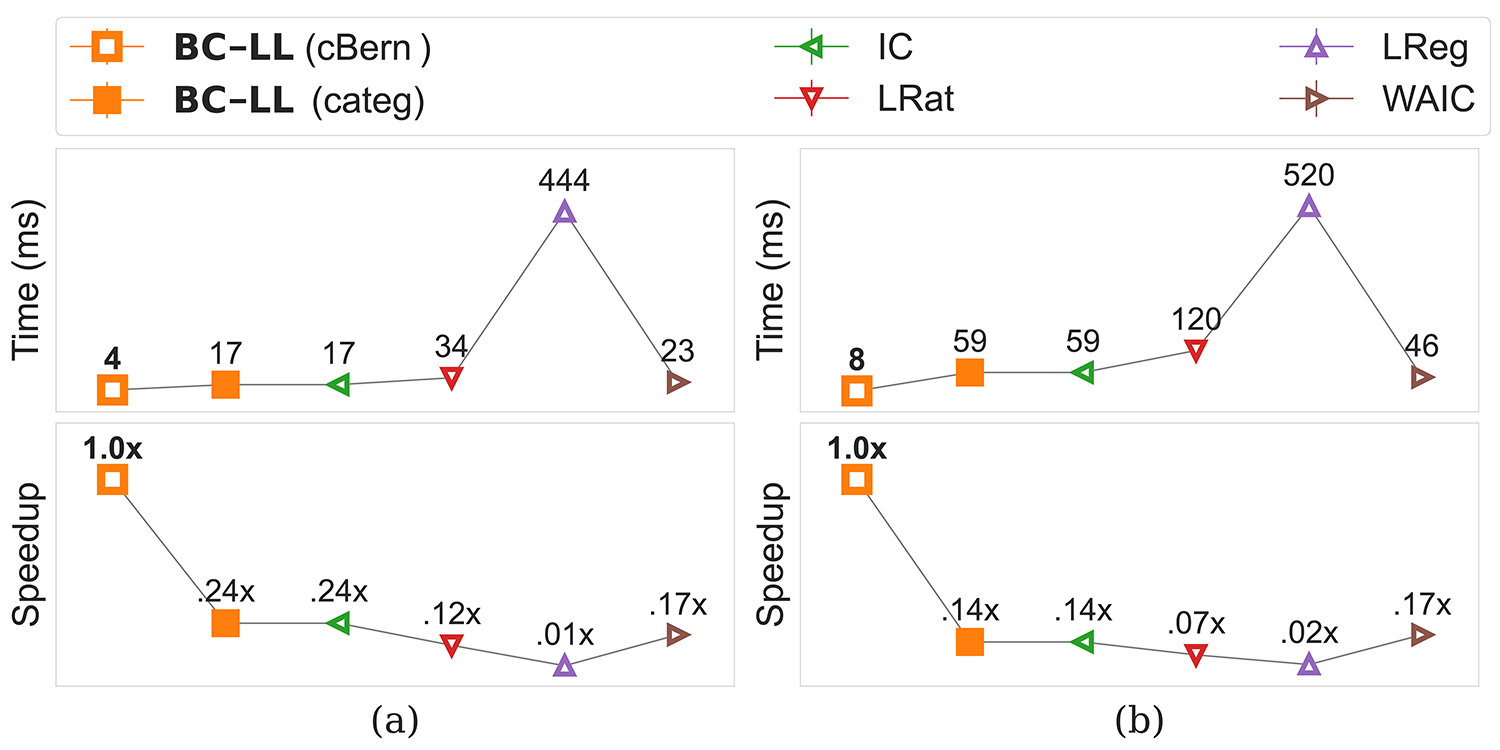}
\caption{{\bf Bias correction is significantly faster than competing approaches.}
\textbf{(a)} (Top row) Compute times (ms) averaged across 500 test samples with each approach, for a grayscale image VAE (lower is better). (Bottom row) Speedup factor computed as the ratio of the compute time for each approach to the compute time for the BC-LL score (higher is better) \textbf{(b)} Same as in panel (a), but showing compute times (top row) and speedups (bottom row) using a natural image VAE. Other conventions are the same in Figures~\ref{fig:grayscale} and~\ref{fig:natural}.}
\label{fig:ct}
\vspace{-1em}
\end{figure}

Finally, we compared compute times for our scores with previous outlier detection approaches, based on the inference time averaged across 500 randomly selected samples from each test dataset. Compute times for BC-LL scores outperformed competing approaches by a factor of up to 100x (Fig.~\ref{fig:ct}), for both grayscale and natural image datasets. In summary, the bias corrected likelihood provides a simple, and computationally inexpensive approach for achieving state-of-the-art outlier detection.

\section{Limitations and Conclusions}
We have proposed, and extensively tested, a simple and efficient remedy to achieve robust outlier detection by de-biasing VAE likelihoods. We demonstrate the effectiveness of this remedy for outlier detection with multiple grayscale and natural image datasets (total of 9 datasets), using four different, popular VAE visible distributions (continuous Bernoulli, Bernoulli, categorical and truncated Gaussian), and comparing performance with four competing state-of-the-art, approaches~\cite{Serra2019, Ren2019, Xiao2020, Choi2018}. In nearly all cases, our correction approached or achieved state-of-the-art outlier detection performance. We also show that outlier detection is robust to the number of latent dimensions in the VAE (Appendix E.1). In addition, our remedy is computationally inexpensive and significantly outperforms all competing approaches in terms of evaluation times. In sum, lightweight remedies suffice to ameliorate biases arising from low-level image statistics (e.g. intensity or contrast variations), to achieve robust outlier detection with VAE likelihoods.

Yet, we note that certain kinds of datasets pose a challenge for outlier detection with VAE likelihoods. For example, even with bias correction VAE likelihoods failed to achieve superlative outlier detection with the CIFAR-10 dataset (Fig.~\ref{fig:natural}, last column). In fact,  nearly all of the other competing approaches under-performed when tested with this dataset. These results suggest that the failure may be due to the unique nature of the CIFAR-10 dataset. Should we, or should we not, then discount VAE likelihoods as a generally applicable metric for outlier detection?

To answer this question we ask what, specifically, the VAE ELBO objective seeks to optimize. On the one hand, the VAE must learn a sparse representation of features in a low-dimensional latent space (the KL-divergence term). On the other hand, the VAE must also faithfully reconstruct the image, pixel-for-pixel (negative reconstruction error). In fact, minimizing the reconstruction error is key to achieving high VAE likelihoods. To achieve this latter objective, the VAE must also learn spatial relationships among its input features in a coordinate frame locked to the bounding frame (edges) of the image. Consequently, VAE likelihoods cannot (and do not) reflect solely high-level, semantic features of the image. 

We propose that VAE likelihoods are ideally suited for outlier detection for datasets with relatively homogeneous images, in which features of the foreground objects (e.g. faces, cars) appear at consistent spatial locations in a coordinate frame locked to the edge of the image (e.g. CompCars, GTSRB, CelebA, SVHN). In all of these cases, our bias correction sufficed to achieve excellent, and state-of-the-art outlier detection. Yet, when image features associated with the foreground object are heterogeneously and inconsistently distributed relative to edge coordinates (e.g. CIFAR-10), bias corrected likelihoods were not as effective. This reasoning also explains why  conventional deep network models~\cite{Lakshminarayanan2017}, or approaches based on optimizing VAE latent feature representations~\cite{Xiao2020}, or approaches that train models on deep classifier-based semantic feature representations~\cite{Krichenko2020} may outperform conventional VAE likelihood-based approaches for outlier detection, especially with  heterogeneous datasets like CIFAR-10. We explore this hypothesis further in Appendix G.

Nonetheless, the bias correction we propose for VAEs will be relevant for outlier detection in key real-world applications. For example, medical imaging data such as retinal scans or MRI images are typically acquired in highly stereotyped spatial coordinates. In these cases, VAEs can be deployed as expert deep learning systems that are sensitive to fine-grained spatial relationships among image features (e.g. the morphology of pathological tissues) for accurate outlier detection. Future work will explore these applications.

{\small
\bibliographystyle{ieee_fullname}
\bibliography{cvpr}
}

\clearpage

\appendix
\section*{Appendices: Robust outlier detection by de-biasing VAE likelihoods}


\section*{Appendix A: VAE architecture and training}
\noindent All experiments were performed using Tensorflow 2 and Tensorflow Probability libraries. We employed a convolutional VAE architecture that follows the DCGAN~\cite{DCGAN2016}  structure (Table \ref{tab:vae_arch}), nearly identical with that of ~\cite{Xiao2020}. We used the Adam optimizer~\cite{ADAM2015} with a learning rate of 5e-4 for training all of our models. Each model was trained for 1000 epochs with a batch size of 64, and the checkpoint with best validation performance based on negative log likelihood was used for reporting results. We used the Xavier uniform initializer (default in Tensorflow 2) for initializing  network weights.

For reporting results based on the bias-corrected log likelihood (BC-LL score) we used a VAE with a latent dimension (nz) of size 20. To examine the robustness of our remedies to the VAE architecture, we also trained four additional VAEs with latent dimensions of size 40, 60, 80, and 100 (see Appendix E.1, Fig.~\ref{fig:roc_zdim}). The same architecture was used for training both grayscale and natural image VAEs, with two differences (grayscale: nf = 32, nc = 1; natural: nf = 64, nc = 3). Log likelihoods were estimated using the importance weighted lower bound (n=100 samples)~\cite{IWAE}.

\begin{table}[h]
\caption{{\bf VAE architecture}. nc: number of channels; nf: number of filters; nz: number of latent dimensions; BN: batch normalization; Conv: convolution layer; DeConv: deconvolution layer; ReLU: rectified linear unit}
\centering
\resizebox{\linewidth}{!}{%
\begin{tabular}{ll}
\toprule
\textbf{Encoder} & \textbf{Decoder} \\

\midrule
Input image of shape 32 $\times$ 32 $\times$ nc & Input latent code, reshape to  1 $\times$ 1 $\times$ nz\\
4 $\times$ 4 Conv\textsubscript{nf} Stride=2, BN, ReLU & 4 $\times$ 4 DeConv\textsubscript{4 $\times$ nf} Stride=1, BN, ReLU \\
4 $\times$ 4 Conv\textsubscript{2 $\times$ nf} Stride=2, BN, ReLU & 4 $\times$ 4 DeConv\textsubscript{2 $\times$ nf} Stride=2, BN, ReLU \\
4 $\times$ 4 Conv\textsubscript{4 $\times$ nf} Stride=2, BN, ReLU & 4 $\times$ 4 DeConv\textsubscript{ nf} Stride=2, BN, ReLU \\
4 $\times$ 4 Conv\textsubscript{2 $\times$ nz} Stride=1 & 4 $\times$ 4 DeConv\textsubscript{nc} Stride=2 \\
\bottomrule
\end{tabular}
}

\label{tab:vae_arch}
\end{table}

\section*{Appendix B: Bias correction for Bernoulli decoder}
For a VAE decoder with a Bernoulli visible distribution, the negative reconstruction error is given by: 
\begin{align*}
\log p_\theta (\mathbf{x} | \mathbf{z}) &= \log p_{\textsc{b}}(\mathbf{x}; \mathbf{\hat{x}}_\theta(\mathbf{z})) \\ &= \sum_{i=1}^D x_i \log \hat{x}_i + (1-x_i) \log (1-\hat{x}_i)
\end{align*}
where $x_i$ is the pixel value of the $i^\mathrm{th}$ pixel in the input sample and $\hat{x}_i$ (or $\hat{x}_i(\mathbf{z})$) is the corresponding pixel value in the image reconstructed by the decoder, and $\mathbf{z}$ is the latent representation corresponding to the input image (see~\cite{Kingma2013} their Appendix C.1).

The negative reconstruction error for perfect reconstruction is simply calculated by setting $\hat{x}_i = x_i$, as: 
\begin{align*}
\log p_{\textsc{b}}(\mathbf{x}; \mathbf{x}) &= \sum_{i=1}^D x_i \log x_i + (1-x_i) \log (1-x_i)
\end{align*}

\section*{Appendix C: Data sources and pre-processing}

\textbf{Data sources}. We used Tensorflow Datasets\footnote{\url{https://www.tensorflow.org/datasets}} for all datasets except Sign Language MNIST, CompCars and GTSRB. We fetched the Sign Language MNIST\footnote{\url{https://www.kaggle.com/datamunge/sign-language-mnist}},  CompCars\footnote{\url{http://mmlab.ie.cuhk.edu.hk/datasets/comp_cars/}} surveillance-nature images and GTSRB\footnote{\url{https://benchmark.ini.rub.de/gtsrb_dataset.html}} datasets from their respective official sources.  For the EMNIST dataset, we selected only the ``Letters'' split. To generate noise patches, we used uniform random noise in range [0, 1], sampled independently across pixels and channels. 
\\
\\
\noindent 
\textbf{Data pre-processing and splits} For both grayscale and natural image datasets, we resized all images to 32 $\times$ 32 pixels before VAE training and evaluation. For natural image datasets, we employed the cropped versions of SVHN and CelebA (default option in Tensorflow Datasets), such that the central object (numbers or faces, respectively) were approximately centered in each image. In addition, we preprocessed the images using ``contrast stretching'', where specified, with the method described in Section~\ref{sec:cs}. For all datasets, we reserved 10\% for the training data for validation (generated twice for two train-val splits, independently). Evaluation was performed with the test splits of each dataset, as indicated in their respective sources. The specific datasets and the number of training and testing samples, and representative exemplars are shown in Table~\ref{tab:data_details}.

\begin{table*}[h!]
\caption{{\bf Dataset details.} Details of datasets used for outlier detection with VAE likelihoods.}
\centering
\resizebox{0.5\pdfpageheight}{!}{%
\begin{tabular}{llcrrl}
\toprule
{\bf Dataset} & {\bf Type} & {\bf Exemplars} & {\bf N-train (N-val)} & {\bf N-test} & {\bf License} \\

\midrule
MNIST & Grayscale & \includegraphics[height=0.35in]{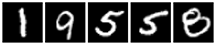} & 54000 (6000) & 10000 & CC BY-SA 3.0 \\
Fashion-MNIST & Grayscale & \includegraphics[height=0.35in]{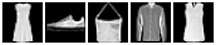} & 54000 (6000) & 10000 & MIT \\
EMNIST-Letters & Grayscale & \includegraphics[height=0.35in]{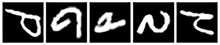} & 79920 (8880) & 14800 & CC BY-SA 3.0\\
Sign Language MNIST & Grayscale & \includegraphics[height=0.35in]{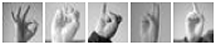} & 24720 (2735) & 7172 & CC0: Public Domain \\
Gray-Noise & Grayscale & \includegraphics[height=0.35in]{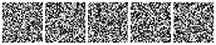} & - & 10000 & - \\

\midrule

SVHN & Color & \includegraphics[height=0.35in]{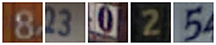} & 65932 (7325) & 26032 & Custom (non-commercial) $^1$\\
CelebA & Color & \includegraphics[height=0.35in]{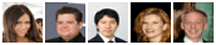} & 146493 (16277) & 19962 & Custom (non-commercial) $^2$\\
CompCars & Color & \includegraphics[height=0.35in]{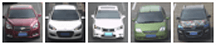} & 28034 (3114) & 13333 & Custom (non-commercial) $^3$\\
GTSRB & Color & \includegraphics[height=0.35in]{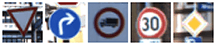} & 35289 (3920) & 12630 & CC0: Public Domain\\
CIFAR-10 & Color & \includegraphics[height=0.35in]{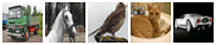} & 45000 (5000) & 10000 & MIT\\
Color-Noise & Color & \includegraphics[height=0.35in]{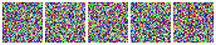} & - & 10000 & - \\

\bottomrule
\\
\multicolumn{5}{l}{$^1$\url{http://ufldl.stanford.edu/housenumbers/}}\\
\multicolumn{5}{l}{$^2$\url{http://mmlab.ie.cuhk.edu.hk/projects/CelebA.html}} \\
\multicolumn{5}{l}{$^3$\url{http://mmlab.ie.cuhk.edu.hk/datasets/comp_cars/}}

\end{tabular}
}
\label{tab:data_details}
\end{table*}

\section*{Appendix D: Performance metrics and competing methods}
\noindent
{\bf Computing performance metrics}. We computed three performance metrics for outlier detection~\cite{Ren2019}: i) area under the receiver-operating-characteristic (AUROC), which represents the area under the curve obtained by plotting the true-positive rate versus the false-positive rate; ii) area under the precision-recall curve (AUPRC), which similarly represents the area under the curve obtained by plotting the precision versus the recall values, and iii) the false-positive rate at 80\% true positive rate (FPR@80\%TPR). Higher values of the first two metrics indicate better outlier detection, and vice versa for the third metric. All metrics were computed with  the {\it scikit-learn} library.

For Figures \ref{fig:grayscale} and \ref{fig:natural}, as well as supplementary results reported in Appendix E, we computed these metrics with an all-versus-all comparison, by comparing each score distribution (e.g. BC-LL) for the test split of the inlier dataset against the respective score distribution for the test split of the outlier dataset (test versus test). 


\noindent
{\bf Implementing competing methods.} 
For implementing competing methods, we used the same VAE architecture as in our paper, except that the number of hidden dimensions was set to 100 to match the likelihood regret paper~\cite{Xiao2020}. Other details are specified below.

\begin{itemize}[leftmargin=*]
    \item {\it Input complexity (IC)}. We computed input complexity by subtracting from the vanilla (uncorrected) log likelihood the ``complexity'' estimate $L(\mathbf{x})$ for each sample computed as $|C(\mathbf{x})|/d$, where the string of bits $C(\mathbf{x})$ was obtained with the PNG compressor ($d$ is the dimensionality of $\mathbf{x}$). Log likelihoods were computed with the categorical visible distribution.
    
    \item {\it Likelihood ratio (LRat)}. We trained a standard VAE and a background model using the noise-corruption procedure described in ~\cite{Ren2019} with the mutation factor set to $\mu$ = 0.3 for grayscale image VAEs and $\mu$ = 0.1 for natural image VAEs. We also applied a large weight decay ($\lambda$ = 100), as suggested by the authors. Log likelihoods were computed with the categorical visible distribution, as in the original study.
    
    \item {\it Likelihood regret (LReg)}. We computed the likelihood regret score by quantifying the improvement in the likelihood for each sample by retraining the encoder for 100 epochs using the implementation provided by the authors~\cite{Xiao2020}. Log likelihoods were computed with the categorical visible distribution, as in the original study.
    
    \item {\it Watanabe-Akaike Information Criterion (WAIC)}. WAIC was computed with the formula $\mathbb{E}_\theta [\log p_\theta (\mathbf{x}) ] - \mathrm{Var}_\theta [\log p_\theta (\mathbf{x}) ]$, using the average log likelihood and variance across the ensemble of six VAEs. Log likelihoods were computed with a continuous Bernoulli visible distribution; the original study~\cite{Choi2018} used a Bernoulli visible distribution.
\end{itemize}

\begin{figure*}[htp]
\centering
   \includegraphics[width=0.9\textwidth]{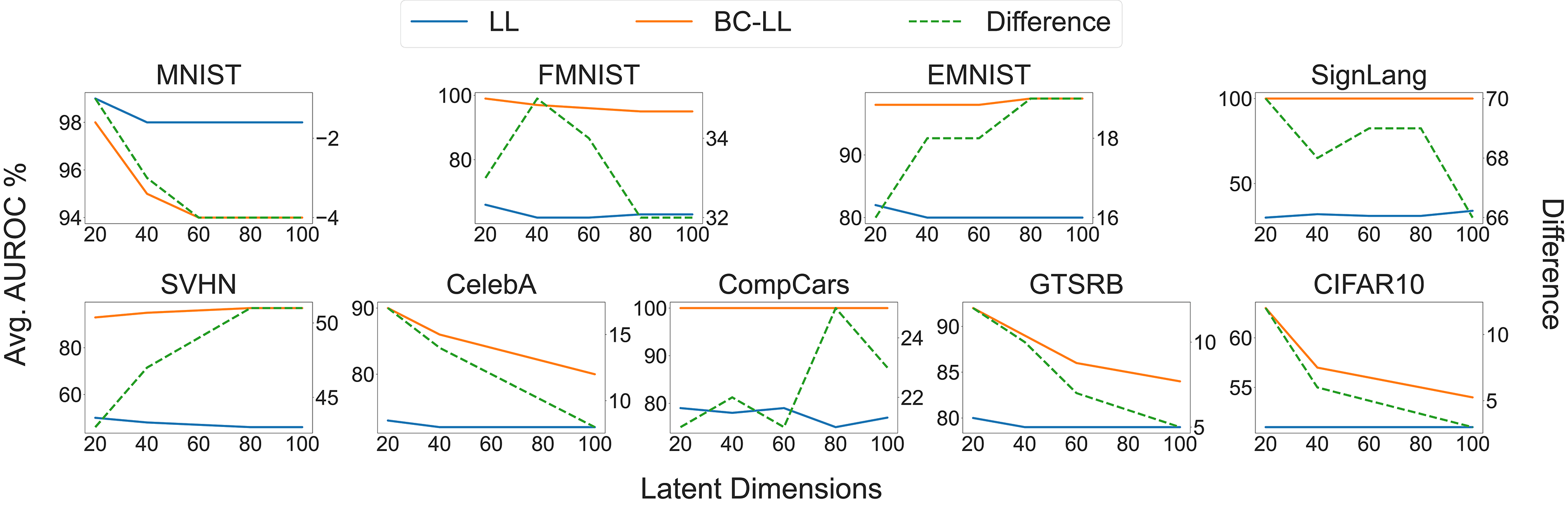}
\caption{{\bf Robustness of outlier detection performance to VAE architecture.}
({\it Top row}) AUROC variation with the number of latent dimensions for the four grayscale image VAEs, based on uncorrected log likelihoods (blue, left y-axis), bias-corrected log likelihoods (orange, left y-axis) and their difference (dashed green, right y-axis). AUROC values were averaged across all (n=4) outlier test datasets.
({\it Bottom row}) Same as in the top panel, but for five natural image VAEs, with AUROC values averaged across n=5 outlier test datasets.
}
\label{fig:roc_zdim}
\end{figure*}

\begin{figure*}[htp]
\centering
\begin{subfigure}{0.45\linewidth}
   \includegraphics[width=\textwidth]{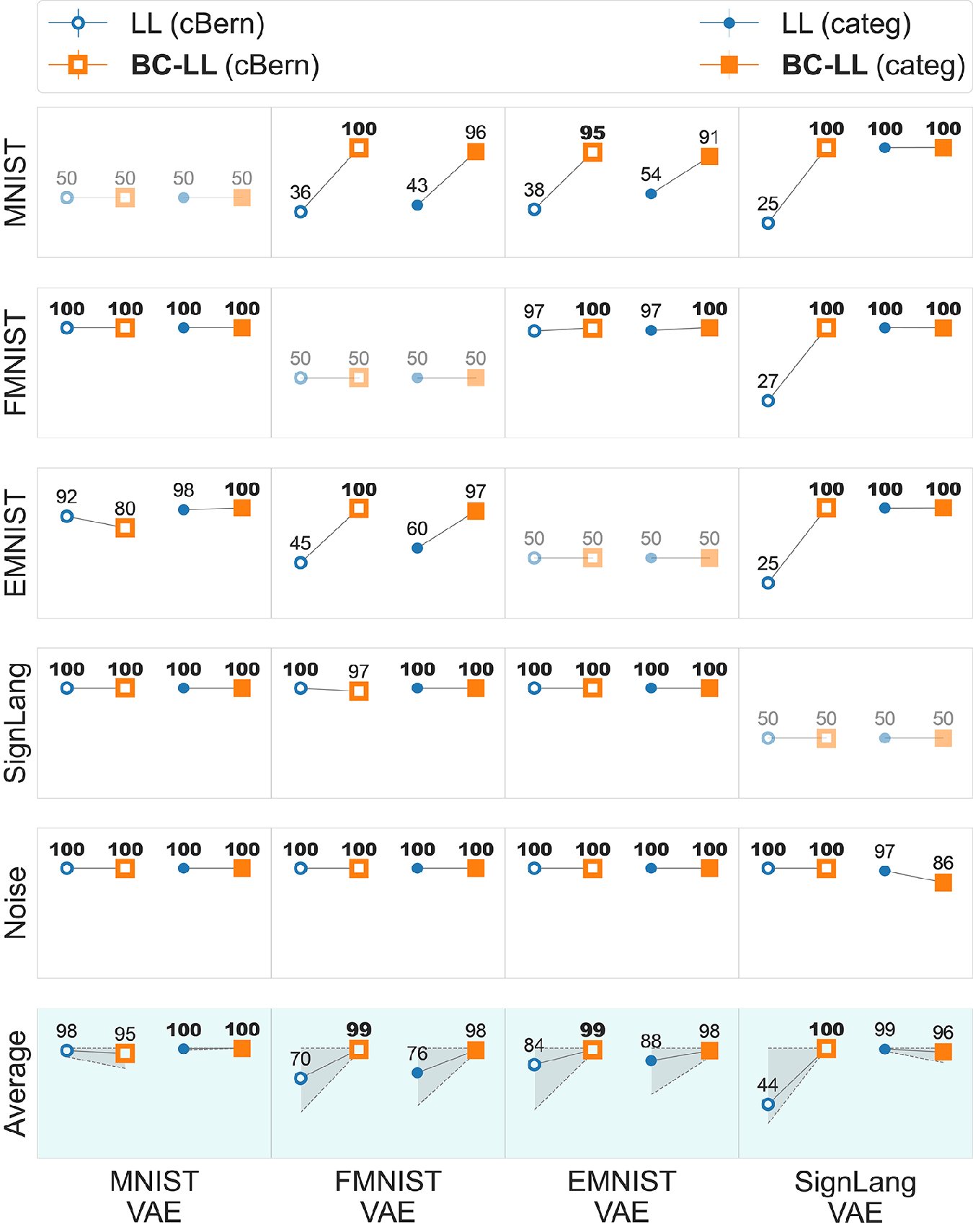}
   \caption{}
   \label{fig:grayscale-prc}
\end{subfigure}
\hspace{2em}
\begin{subfigure}{0.45\linewidth}
   \includegraphics[width=\textwidth]{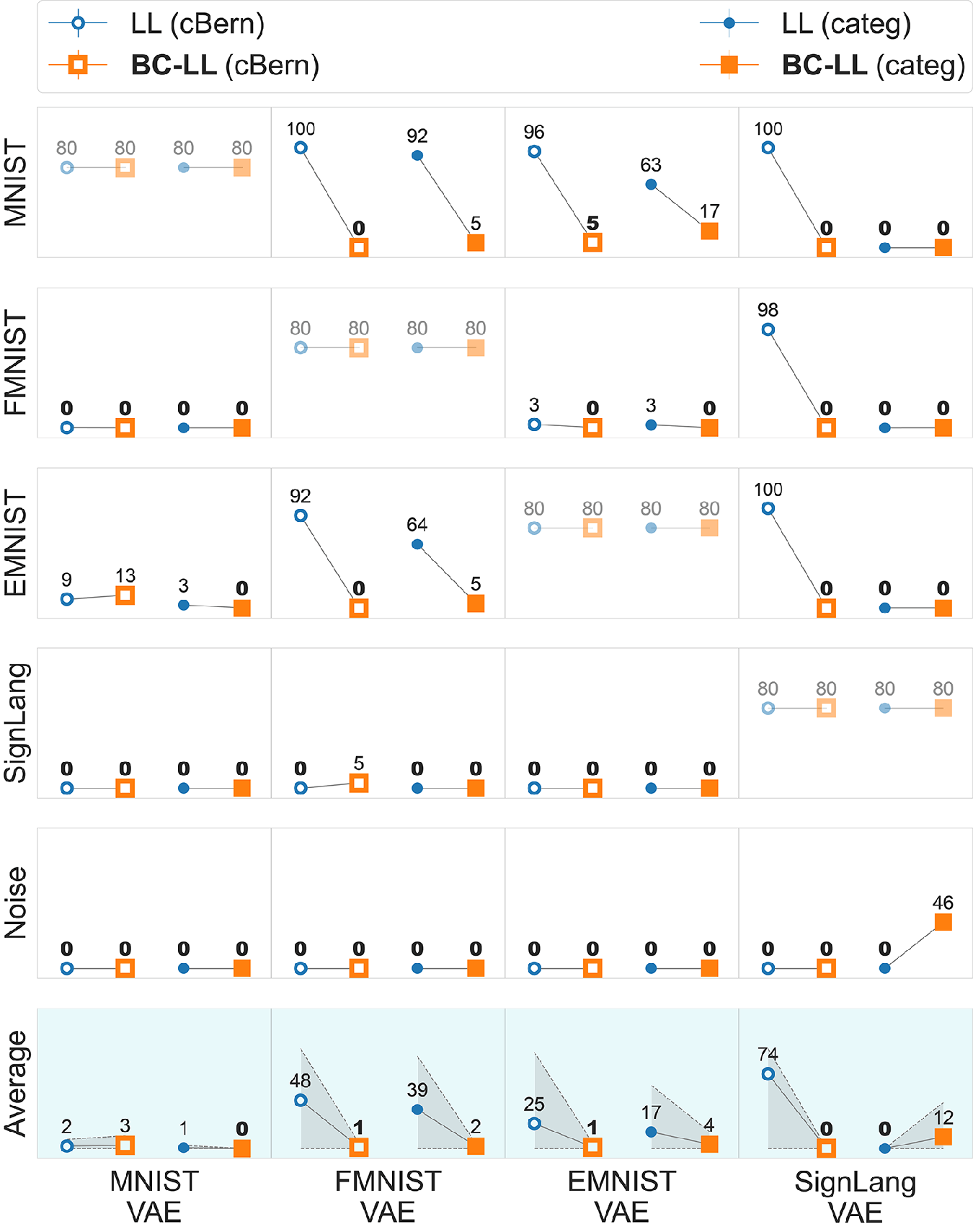}
   \caption{}
   \label{fig:grayscale-fpr}
\vspace{-1em}
\end{subfigure}

\caption{{\bf Additional metrics for outlier detection: Grayscale datasets.} {\bf(a)} Area under the precision recall curve (AUPRC) (higher is better). {\bf(b)} False-positive rate (FPR) at a true-positive rate (TPR) of 80\% (lower is better). Other conventions are the same as in Figure~\ref{fig:grayscale}.} 
\label{fig:grayscale-fpr-prc}
\end{figure*}

\begin{figure*}[htp]
\centering
\begin{subfigure}{0.45\linewidth}
   \includegraphics[width=\textwidth]{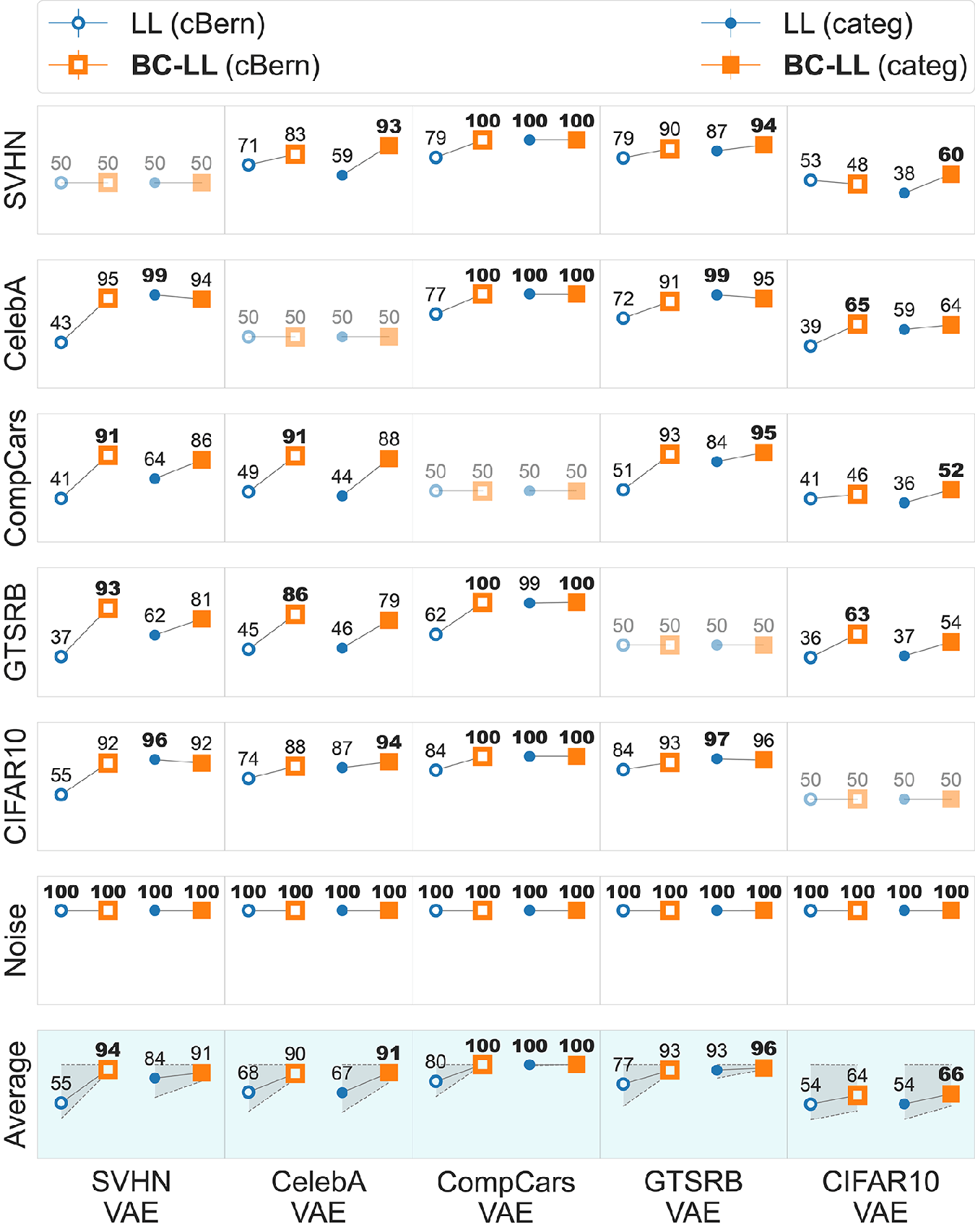}
   \caption{}
   \label{fig:natural-prc}
\end{subfigure}
\hspace{2em}
\begin{subfigure}{0.45\linewidth}
   \includegraphics[width=\textwidth]{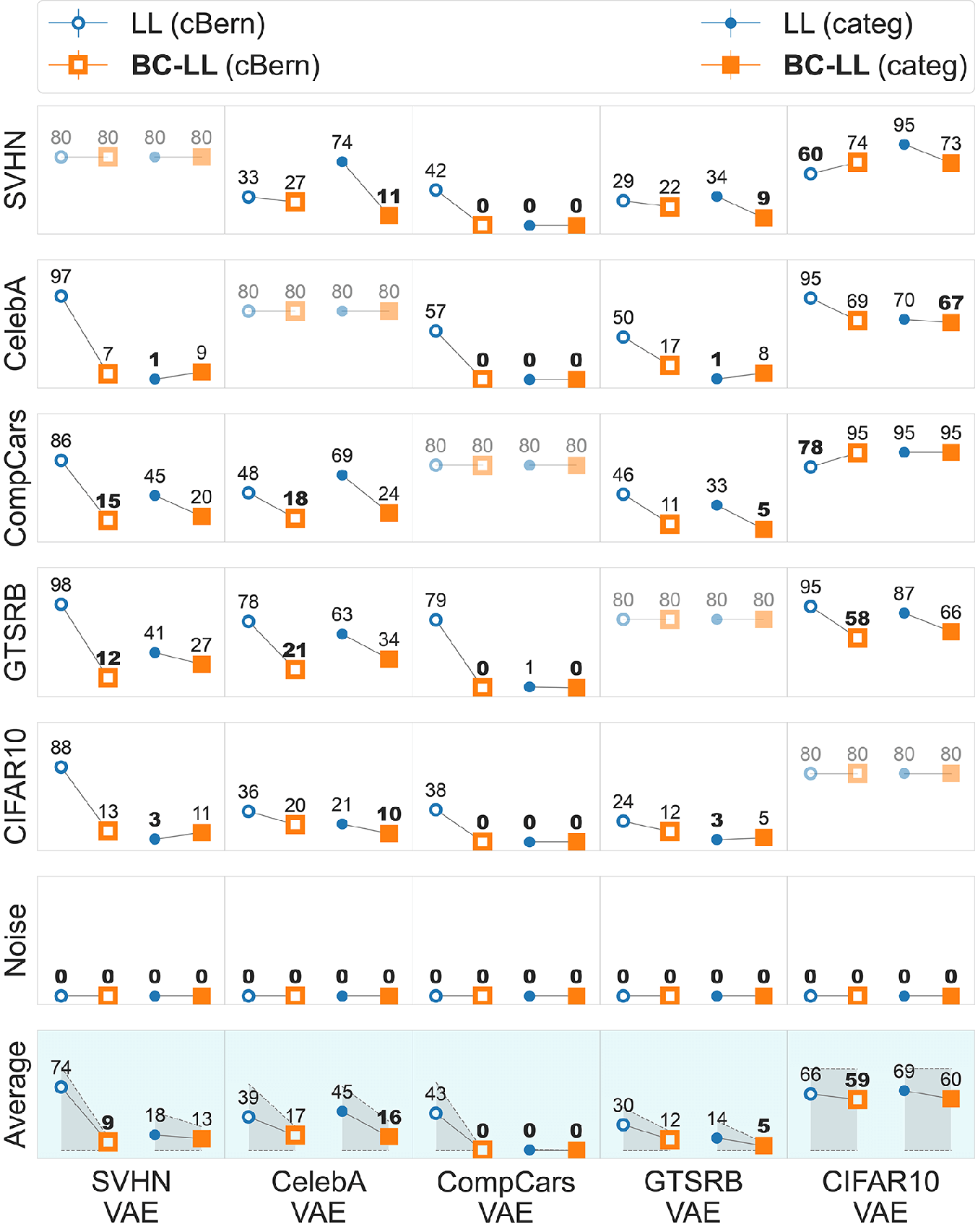}
   \caption{}
   \label{fig:natural-fpr}
\end{subfigure}

\caption{{\bf Additional metrics for outlier detection: Natural image datasets.} Same as in Figures \ref{fig:grayscale-prc} and \ref{fig:grayscale-fpr}, but showing additional metrics for outlier detection, {\bf (a)} AUPRC and {\bf (b)} FPR at 80\%TPR based on natural image VAE likelihoods. Other conventions are the same as in Figure~\ref{fig:grayscale-fpr-prc}.} 
\vspace{-1em}
\end{figure*}

For Figures \ref{fig:grayscale} and \ref{fig:natural} that compare the performance of these other scores with our scores, AUROC values were computed by comparing each score distribution for the test split of the inlier dataset against the score distribution for test split of the outlier dataset (test versus test). The VAEs for competing approaches were trained 3 times with different random initializations; we report average AUROC values across 3 runs.

\section*{Appendix E: Supplementary results on outlier detection}

\subsection*{E.1 Outlier detection performance with varying numbers of VAE
latent dimensions}
We computed outlier detection AUROC for VAEs trained with different numbers of latent dimensions (nz). AUROC values remained more or less constant, or declined marginally with nz (Fig.~\ref{fig:roc_zdim}, blue). In addition, in all cases bias correction improved AUROC values (Fig.~\ref{fig:roc_zdim}, orange versus blue).

\subsection*{E.2 Grayscale image datasets: Additional performance metrics}
In addition to AUROC (Fig.~\ref{fig:grayscale}), we computed AUPRC and FPR at 80\% TPR for outlier detection with the grayscale VAEs with the bias corrected likelihoods (BC-LL). AUPRC typically improved (Fig.~\ref{fig:grayscale-prc}) and FPR decreased (Fig.~\ref{fig:grayscale-fpr}) for BC-LL scores (orange symbols) compared to uncorrected likelihoods (blue symbols).




\subsection*{E.3 Natural image datasets: Additional performance metrics}

In addition to AUROC (Fig.~\ref{fig:natural}), we computed AUPRC and FPR at 80\% TPR for outlier detection with the natural image VAEs. AUPRC typically improved (Fig.~\ref{fig:natural-prc}) and FPR decreased (Fig.~\ref{fig:natural-fpr}) for BC-LL scores (orange symbols) compared to uncorrected likelihoods (blue symbols). As with the AUROC metric (Fig.~\ref{fig:natural}), outlier detection was generally poor with the CIFAR10 VAE.

\subsection*{E.4 Effect of contrast normalization on outlier detection}
\begin{figure}[htp]
\centering
\begin{subfigure}{0.45\linewidth}
   \includegraphics[width=\textwidth]{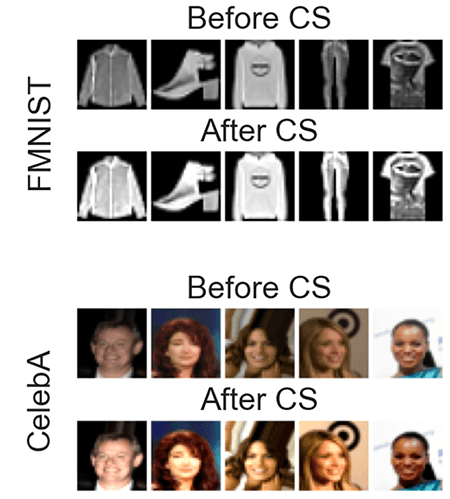}
   \caption{}
   \label{fig:cs_examples}
\end{subfigure}
\hspace{\fill}
\begin{subfigure}{0.50\linewidth}
   \includegraphics[width=\textwidth]{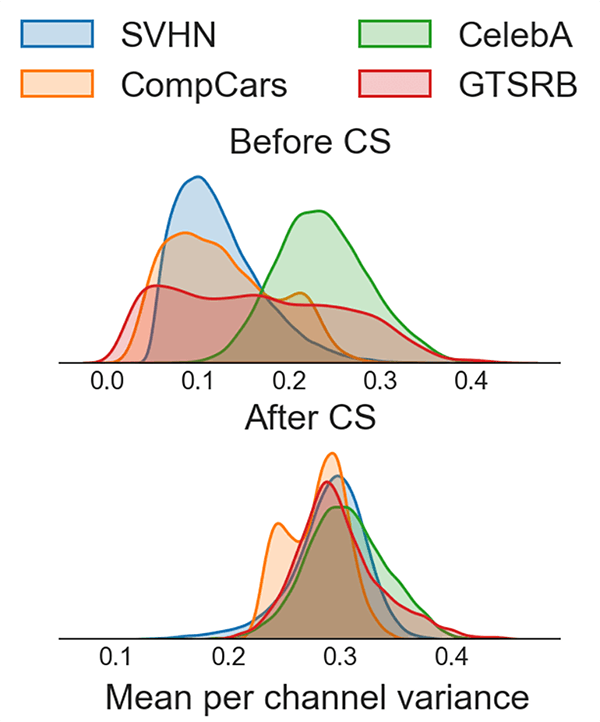}
   \caption{}
   \label{fig:cs_hist}
\end{subfigure}
\caption{{\bf Effect of contrast stretching on image statistics.} {\bf (a)} Representative images from the Fashion-MNIST (top) and CelebA (bottom) datasets before (top row in each sub-panel) and after (bottom row in each sub-panel) contrast normalization (stretching). 
{\bf (b)} Distributions of mean per-channel variance for the natural image datasets show greater overlap after (bottom), as compared to before (top), contrast stretching.}
\vspace{-1em}
\end{figure}

\begin{figure*}[htp]
\centering
\begin{subfigure}{0.42\linewidth}
  \includegraphics[width=\textwidth]{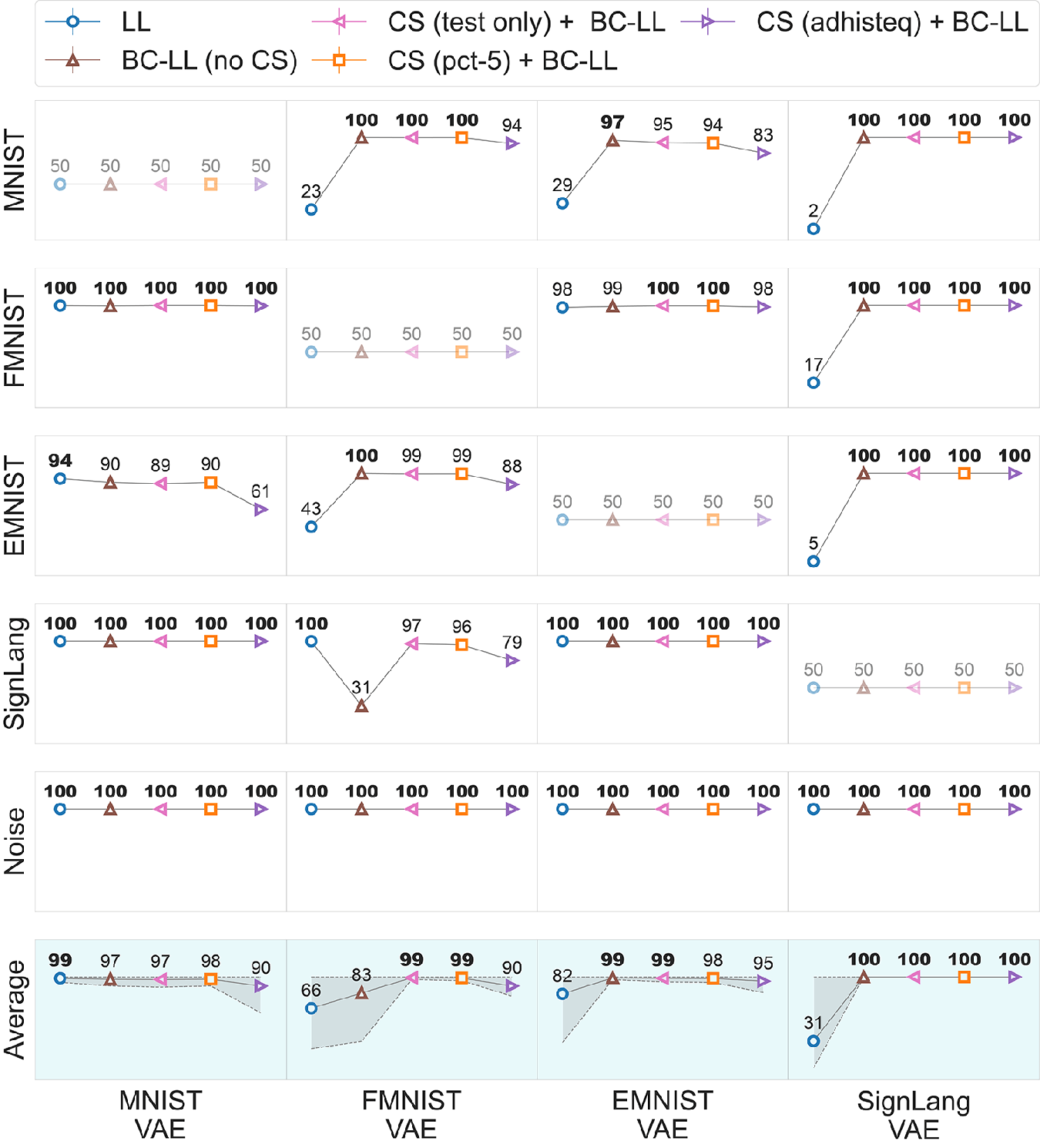}
  \caption{}
\end{subfigure}
\hspace{3em}
\begin{subfigure}{0.42\linewidth}
  \includegraphics[width=\textwidth]{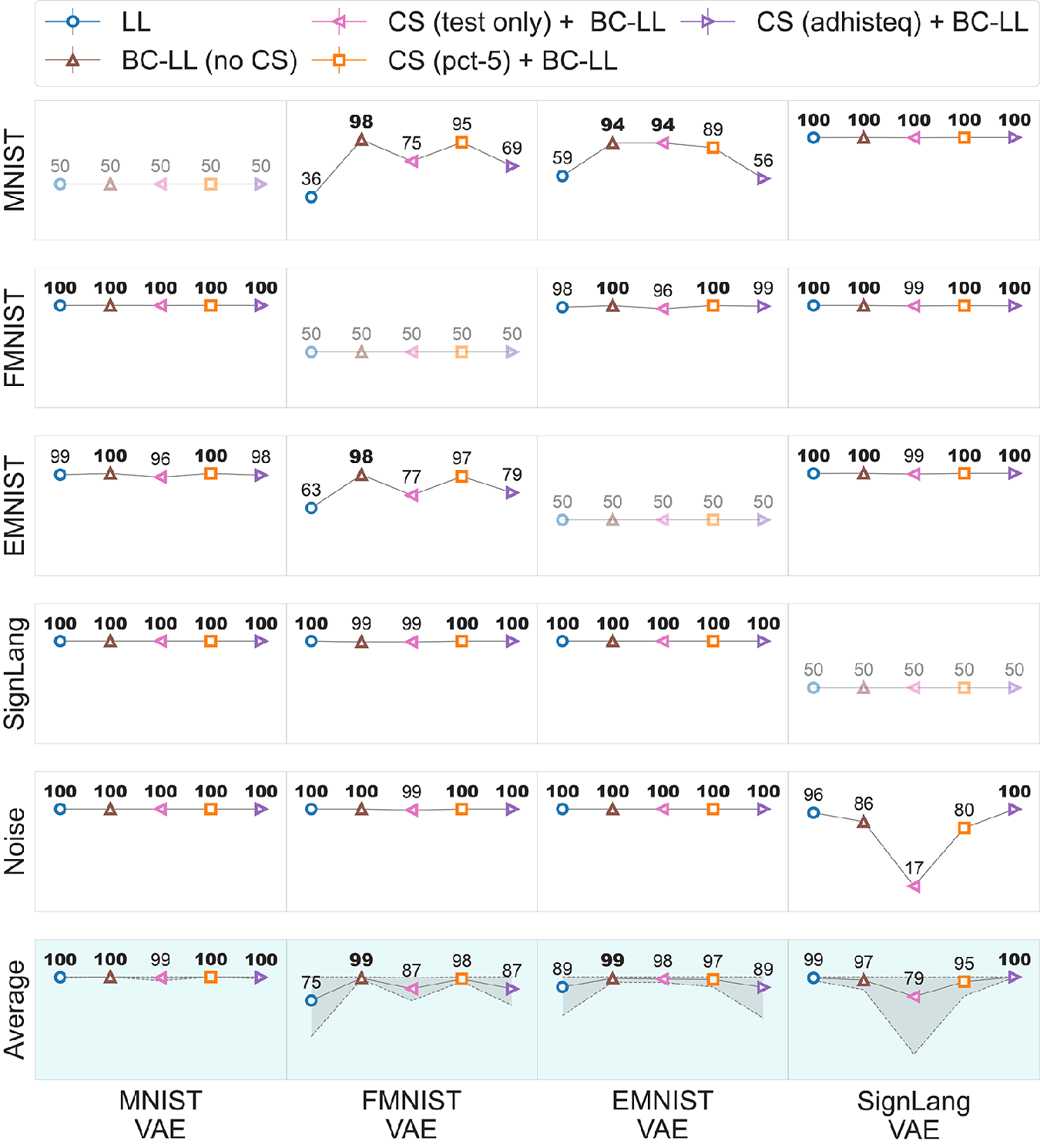}
  \caption{}
\end{subfigure}

\caption{{\bf Effect of contrast normalization on outlier detection: Grayscale datasets.} Outlier detection results using \textbf{(a)} continuous Bernoulli and \textbf{(b)} Categorical VAE likelihoods trained on grayscale image datasets. We report AUROC values computed with no contrast normalization (BC-LL (no CS), brown symbols), with contrast stretching applied only at test time (CS (test only) + BC-LL, pink symbols), with contrast stretching applied at both train and test time (CS (pct-5) + BC-LL) and with adaptive histogram equalization instead of percentile-based contrast stretching (CS (adhisteq) + BC-LL, magenta symbols), all with bias correction. Blue symbols: AUROC based on uncorrected likelihoods. We use adaptive (local) histogram equalization (adhisteq) for grayscale images because global histogram equalization produced unnatural variations in image contrast. Other conventions are the same as in Figure~\ref{fig:grayscale}.} 
\label{fig:grayscale_histeq}
\end{figure*}

\begin{figure*}[htp]
\centering
\begin{subfigure}{0.42\linewidth}
  \includegraphics[width=\textwidth]{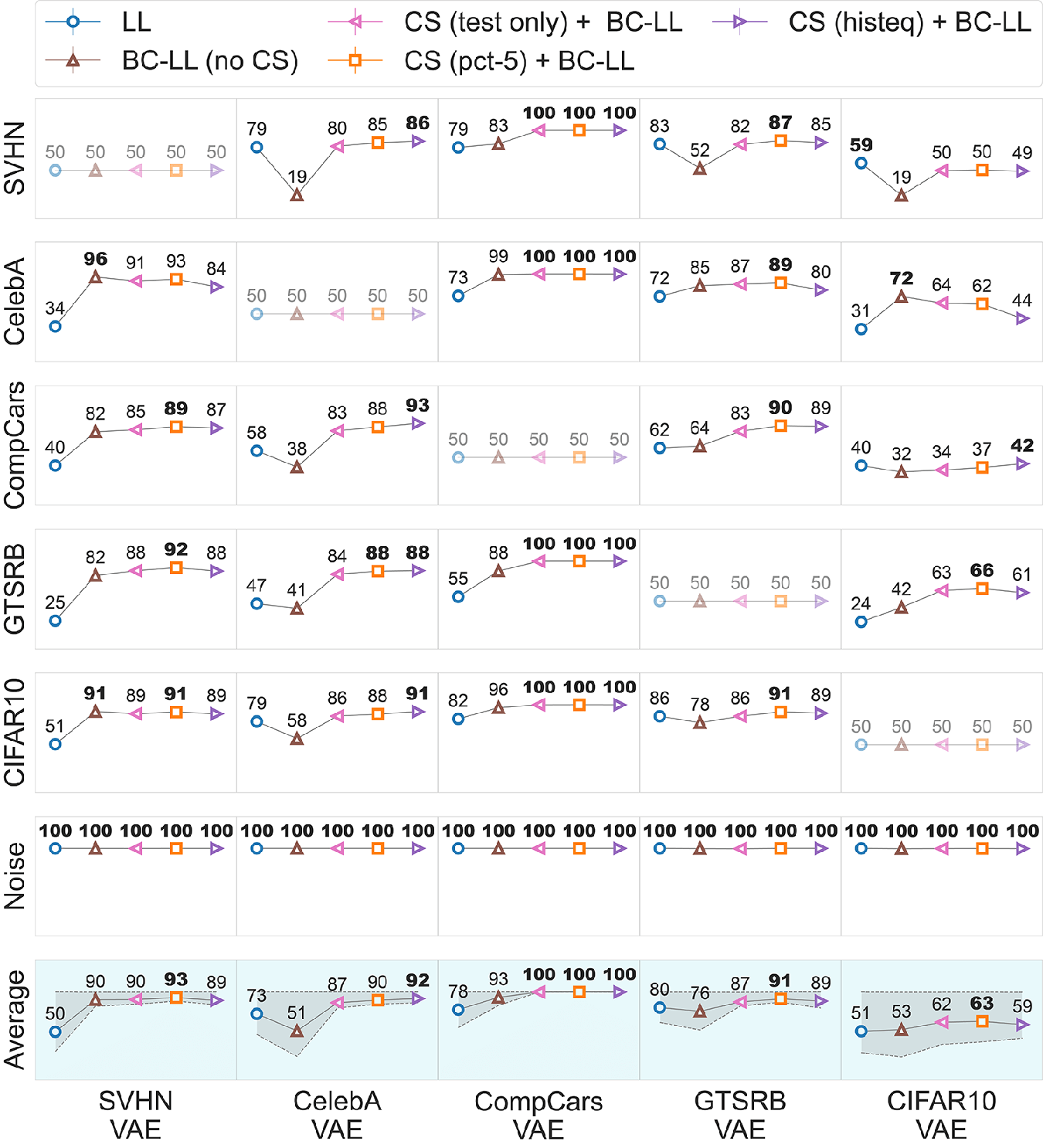}
  \caption{}
\end{subfigure}
\hspace{3em}
\begin{subfigure}{0.42\linewidth}
  \includegraphics[width=\textwidth]{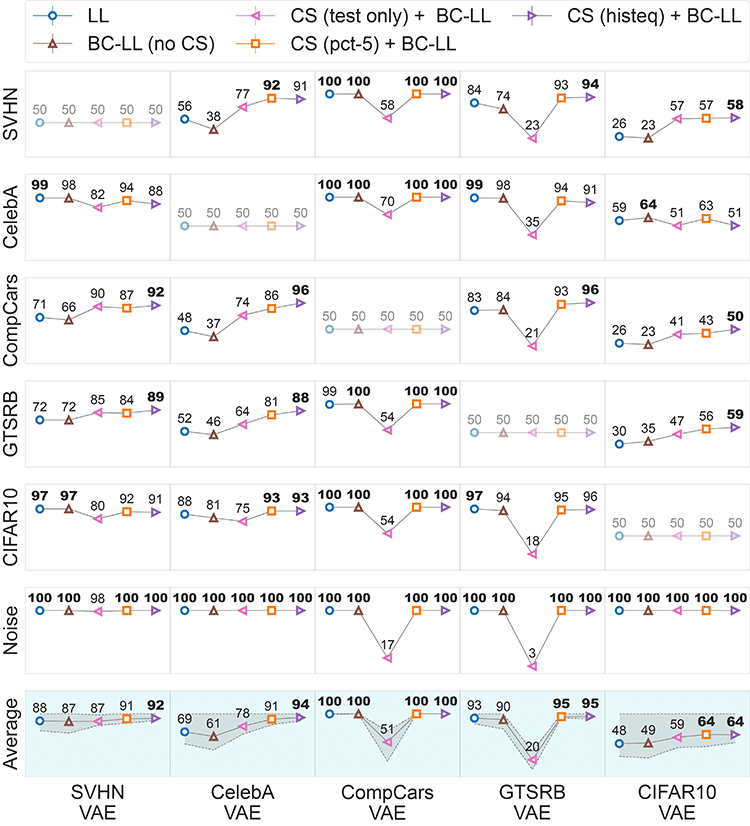}
  \caption{}
\end{subfigure}

\caption{{\bf Effect of contrast normalization on outlier detection: Natural image datasets.} Outlier detection results using \textbf{(a)} continuous Bernoulli and \textbf{(b)} Categorical VAE likelihoods trained on natural image datasets. Magenta symbols (CS (histeq) + BC-LL): AUROC values computed with (global) histogram equalization instead of percentile-based contrast stretching. Other conventions are the same as in Figure~\ref{fig:grayscale_histeq}. } 
\label{fig:natural_histeq}
\vspace{-1em}
\end{figure*}

\begin{figure}[bhtp]
\centering
\begin{subfigure}{1.0\linewidth}
  \includegraphics[width=\textwidth]{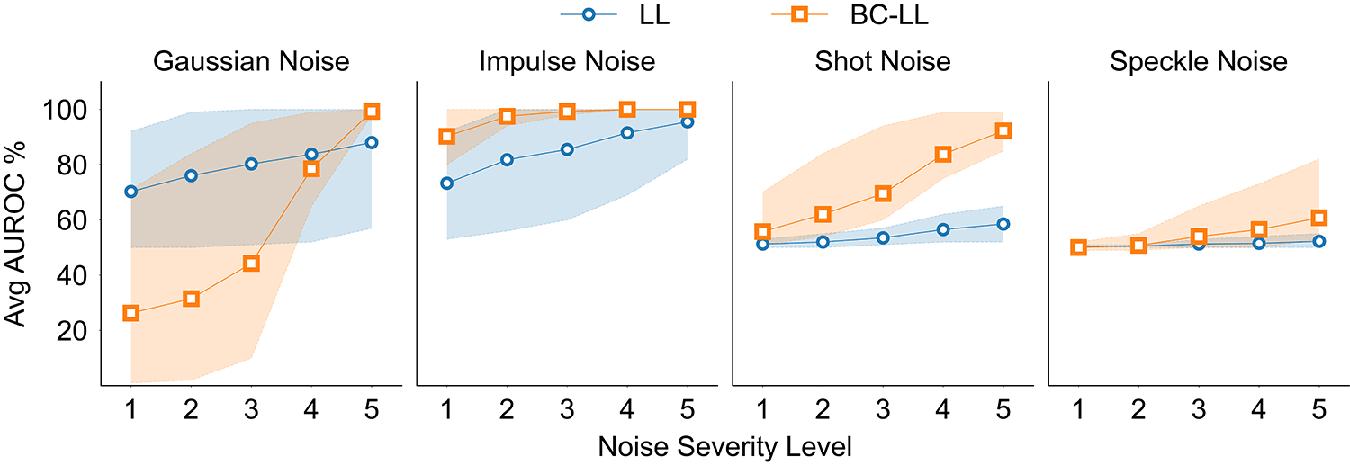}
  \caption{}
  \label{fig:nearood-gs}
\end{subfigure}
\hspace{3em}
\begin{subfigure}{1.0\linewidth}
  \includegraphics[width=\textwidth]{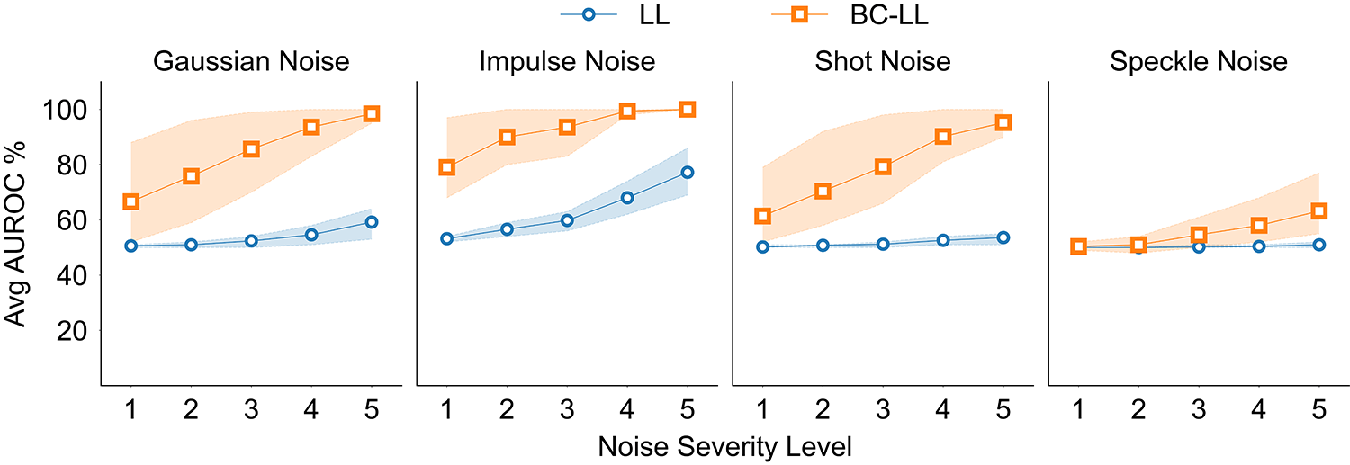}
  \caption{}
    \label{fig:nearood-color}
\end{subfigure}

\caption{{\bf Outlier detection with milder perturbations}: {\bf (a)} AUROC values for near-OOD detection with four different types of noise -- (from left to right) Gaussian, Impulse, Shot and Speckle -- at 5 different levels of severity \cite{Hendrycks2019}. Values averaged across the four grayscale datasets for each noise type and severity level. {\bf (b)} Same as in (a) but near-OOD AUROC values averaged across  the five natural image datasets.}
 
\label{fig:nearood}
\end{figure}

\begin{figure*}[tbh]
\centering
\begin{subfigure}{0.42\linewidth}
  \includegraphics[width=\textwidth]{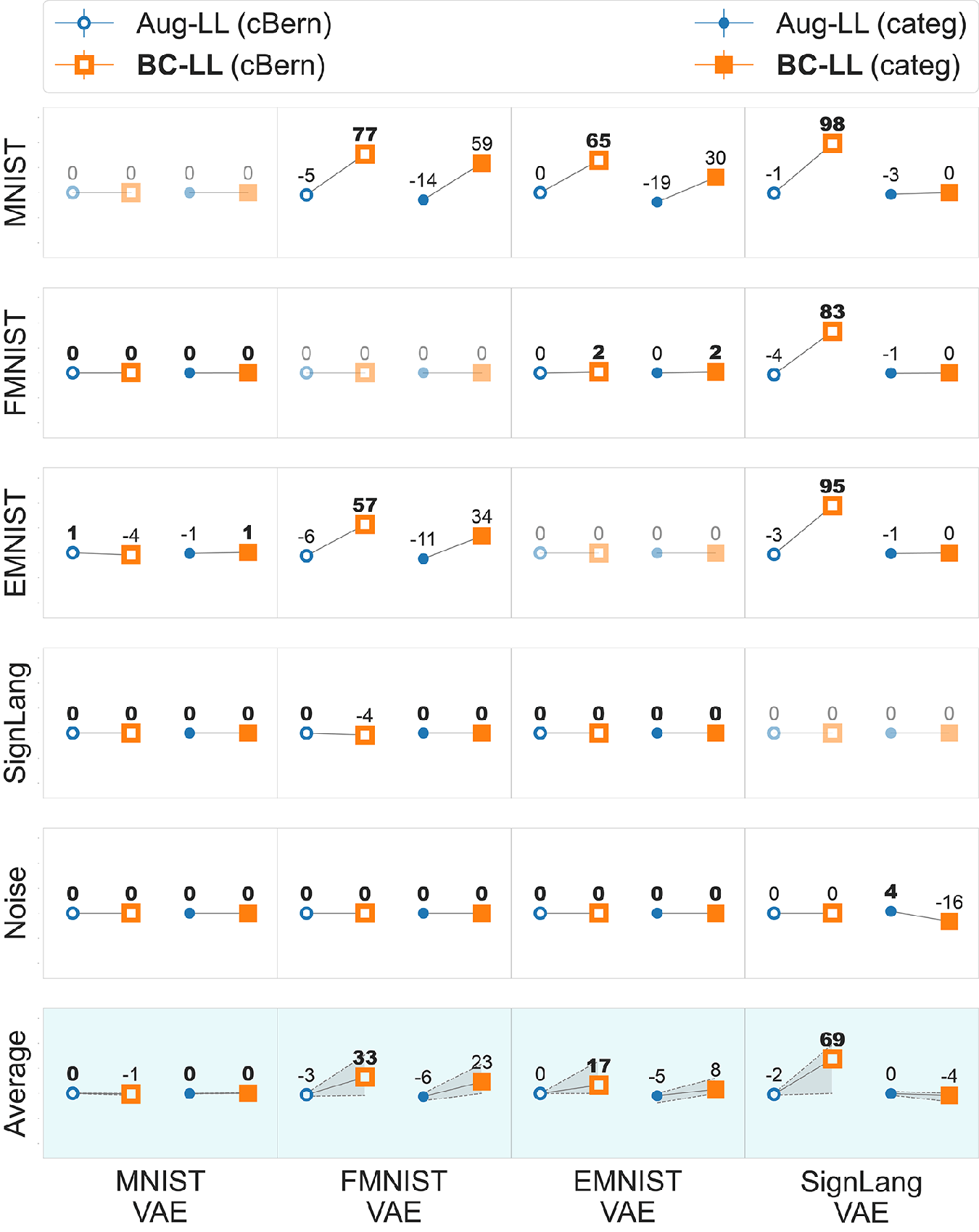}
  \caption{}
\end{subfigure}
\hspace{3em}
\begin{subfigure}{0.42\linewidth}
  \includegraphics[width=\textwidth]{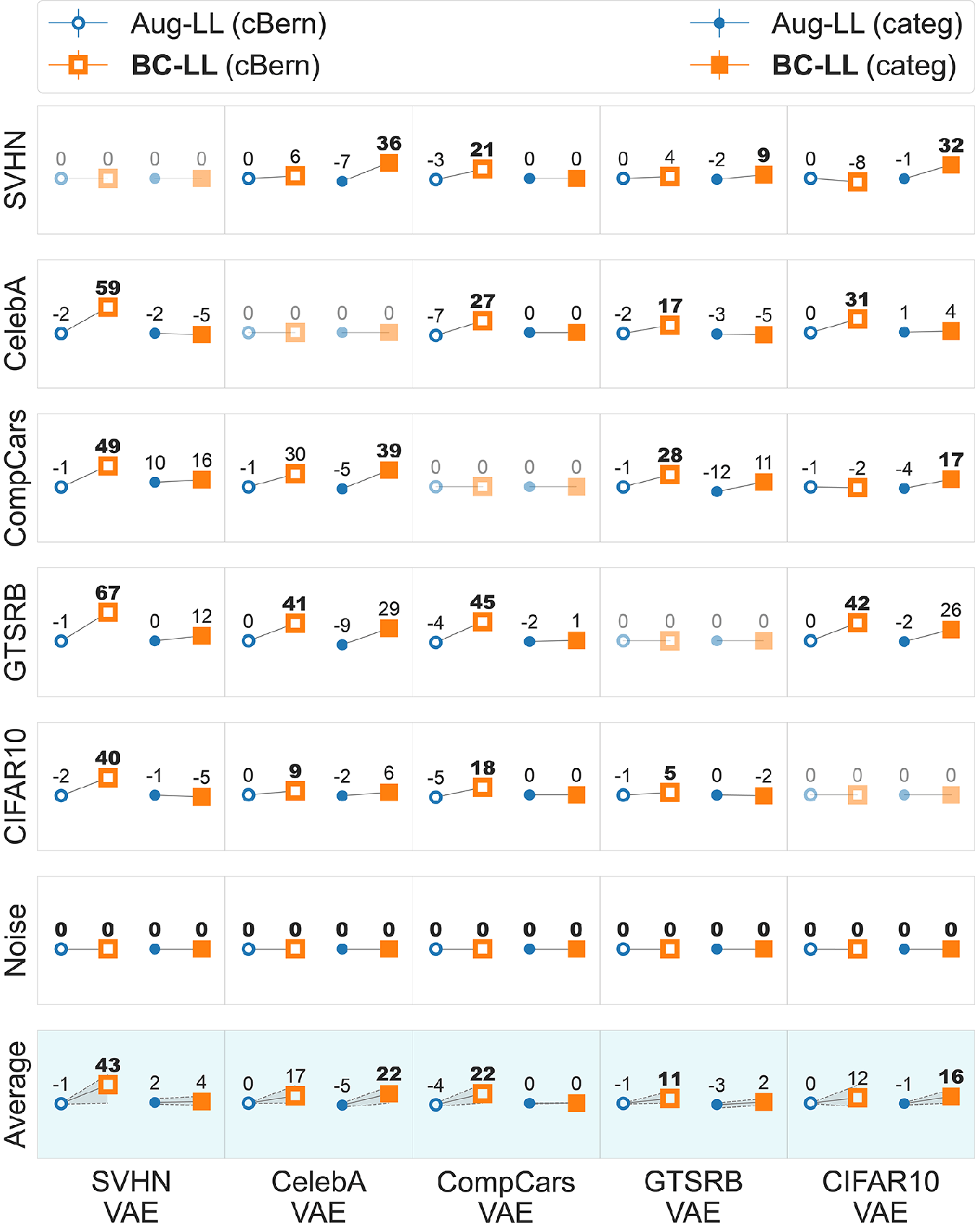}
  \caption{}
\end{subfigure}

\caption{{\bf Augmented training baseline}: {Change in AUROC upon training the VAEs using augmented samples, with varying intensities and contrasts. Change in AUROC was measured relative to the vanilla log likelihoods AUROC  (blue symbols) for (a) grayscale and (b) natural image datasets. Change in AUROC with bias correction (orange symbols) is also shown for each train-test pair, for reference. Other conventions are the same as in Figures~\ref{fig:grayscale} and~\ref{fig:natural}.}}
\label{fig:aug_train}
\end{figure*}

\subsection*{E.5 Outlier detection with milder perturbations}

We tested the efficacy of bias correction to detect milder corruptions to the data (distribution shifts) at test time. We used continuous Bernoulli VAEs trained on multiple grayscale and natural image datasets. At test time, we corrupted ID images with four noise types -- Gaussian noise, impulse noise, shot noise and speckle noise, applied with varying degrees of severity -- following the procedure described in \cite{Hendrycks2019}. We computed AUROC values for near-OOD detection with the original images from the ID test sets as ID samples and their noise-corrupted variants as OOD samples, for both grayscale (Fig.~\ref{fig:nearood-gs}) and natural image  (Fig.~\ref{fig:nearood-color}) datasets. Bias correction (CS(pct-5) + BC-LL) improved outlier detection performance in most cases, typically in a manner that increased with noise severity level.

\subsection*{E.6 Augmented training baseline}
As a baseline, we asked whether training each VAE using augmented samples, with varying contrasts and intensities, would improve outlier detection. We augmented the training samples by varying the pixel intensity uniformly across the image (up to ±75 pixel units), as well as by varying contrast using percentile-based contrast stretching (up to 10\%). We trained continuous Bernoulli and categorical VAEs on the augmented training sets for all of the four grayscale and five natural image datasets. We then computed AUROC values for OOD detection on the original test sets (without augmentation); we report the results as a difference in AUROC values relative to those obtained with vanilla VAE likelihoods. In general, augmented training either did not improve, or marginally worsened, AUROC for outlier detection, relative to the vanilla likelihoods (Fig.~\ref{fig:aug_train}, blue). By contrast, bias correction (CS(pct-5) + BC-LL) yielded significant improvements over vanilla likelihoods in nearly all cases (Fig.~\ref{fig:aug_train}, orange).

\subsection*{E.7 Effect of preprocessing with ZCA whitening}
We experimented with ZCA whitening, as an alternative to percentile based contrast stretching. The ZCA whitening transform was computed and applied for each image, to normalize per channel variance and to decorrelate values across the three color channels. We trained continuous Bernoulli VAEs on ZCA-whitened images and computed outlier detection AUROCs also on whitened images at test time, followed by bias correction. In general, ZCA whitening yielded  outlier detection performance comparable with percentile-based contrast stretching for grayscale images (Table~\ref{tab:grayscale_zca_table}) but worse than contrast stretching for natural image datasets (Table~\ref{tab:color_zca_table}).


\begin{table}[bhtp]
\centering
\caption{ZCA+BC-LL AUROC values for VAEs trained with grayscale image datasets. Columns indicate training datasets (ID) and rows indicate test datasets (OOD). (MN - MNIST, FM - Fashion MNIST, EM - EMNIST, SL - SignLang. MNIST)}
\begin{tabular}{lrrrrrr}
\toprule
 MN &    $50$ &  $100$ &  $97$ &  $100$ \\
FM &   $100$ &    $50$ &   $99$ &  $100$ \\
EM &  $88$ &  $100$ &    $50$ &  $100$ \\
SL &   $100$ &   $42$ &  $100$ &    $50$ \\
 Noise &   $100$ &   $100$ &   $100$ &  $100$ \\
\midrule
$\text{OOD}\uparrow \text{\textbackslash}  \ \text{ID}\rightarrow$ &           MN &          FM &          EM &          SL \\
\bottomrule
\end{tabular}
\label{tab:grayscale_zca_table}
\end{table}





\begin{table}[bhtp]
\centering
\caption{ZCA+BC-LL AUROC values for VAEs trained with natural image datasets. Columns indicate training datasets (ID) and rows indicate test datasets (OOD). (SV - SVHN, CA - CelebA, CC - CompCars, GT - GTSRB, CF - CIFAR10.)}
\begin{tabular}{lrrrrr}
\toprule

      SV & $50$ &       $95$ &   $100$ &  $98$ &    $85$ \\
    CA &  $34$ &    $50$ &   $99$ &  $75$ &   $49$ \\
  CC &  $5$ &  $39$ &    $50$ &  $38$ &   $17$ \\
     GT &  $38$ &    $72$ &   $99$ &    $50$ &    $53$ \\
   CF &  $39$ &  $73$ &   $99$ &  $79$ &     $50$ \\
     Noise &  $100$ &   $99$ &   $100$ &   $100$ &   $100$ \\
\midrule
$\text{OOD}\uparrow \text{\textbackslash} \ \text{ID}\rightarrow$  &            SV &          CA &        CC &           GT &          CF \\

\bottomrule
\end{tabular}

\label{tab:color_zca_table}
\end{table}

\section*{Appendix F: Bias in alternative visible distributions}

\subsection*{F.1 Bias in the Categorical visible distribution}
\begin{figure}[h!]
\centering
   \includegraphics[width=\linewidth]{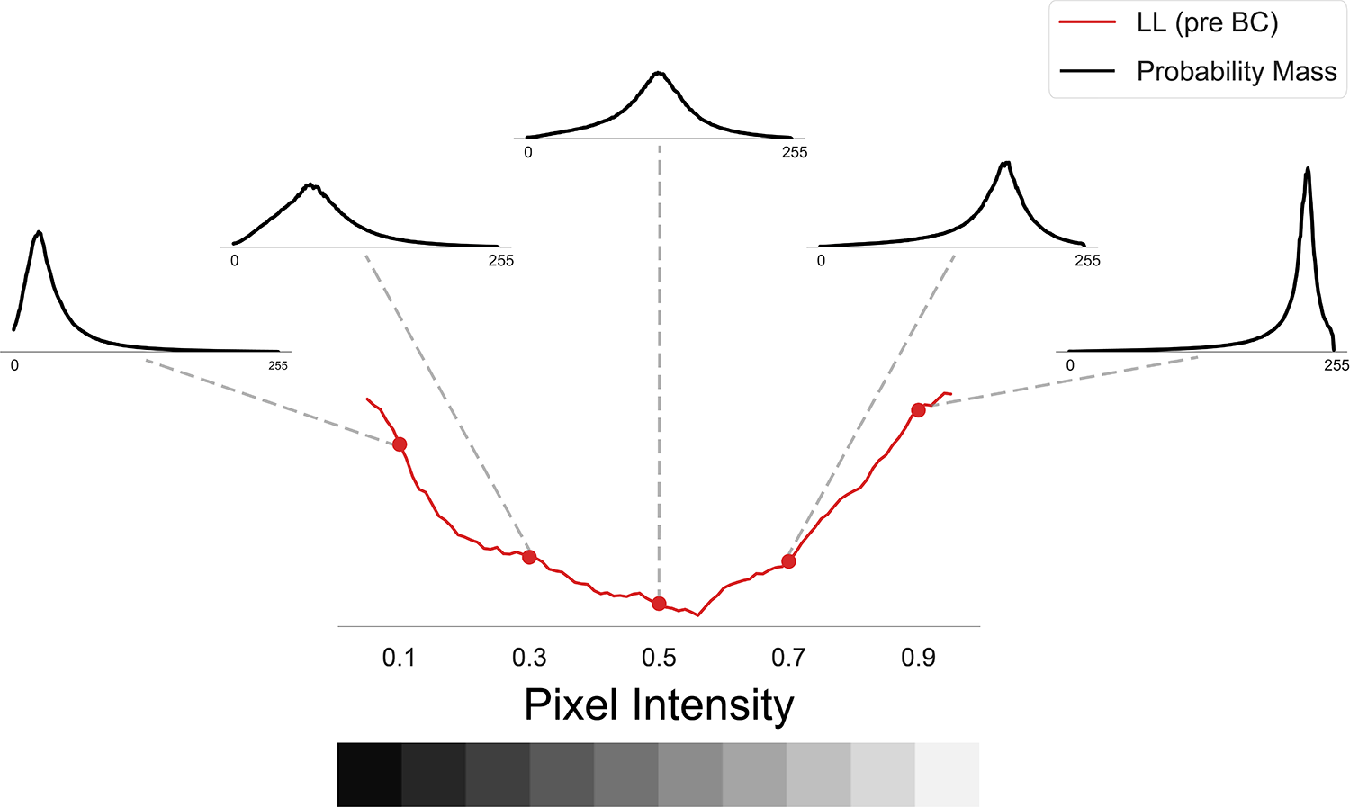}
   \caption{{\bf Empirical bias arising from pixel intensities for a Categorical VAE.} (Red curve) Uncorrected log likelihoods for uniform pixel intensity images. (Insets, black) PMF plots: Categorical distribution outputs for specific target pixel values (from left to right) 25, 76, 127, 178 and 229}
   \label{fig:cat_exp}
\end{figure}

In Figure~\ref{fig:cat_exp}, we revisit the empirical bias in the log likelihood of a CelebA VAE with Categorical visible distribution, for images with different uniform pixel intensities. We plot the VAE decoder categorical distribution outputs for 5 different target pixel intensity values across the range (0-255), by averaging across all occurrences of the respective pixel value in the CelebA test set (Fig.~\ref{fig:cat_exp}). The U-shaped log likelihood profile can be explained by the discrepancy in the entropy of these distributions across different pixel values.  For example, for target pixel values close to full black or full white, the VAE decoder concentrates probability mass in the proximity of the target pixel value, with the output having narrower peaks and smaller overall entropy. For target values closer to the middle of the range (grays), the probability mass is more dispersed around the target pixel value, and has wider peaks and larger overall entropy in the output. These discrepancies in the output entropy across different pixel intensities arise from ``edge effects'': probability mass cannot be assigned to values $<$0 or $>$255, and the PMF has to sum to one for all target pixel outputs, resulting in accumulation of probability mass at the edges.   

\begin{figure*}[ht!]
\centering
\begin{subfigure}[b]{0.2\linewidth}
   \includegraphics[width=\textwidth]{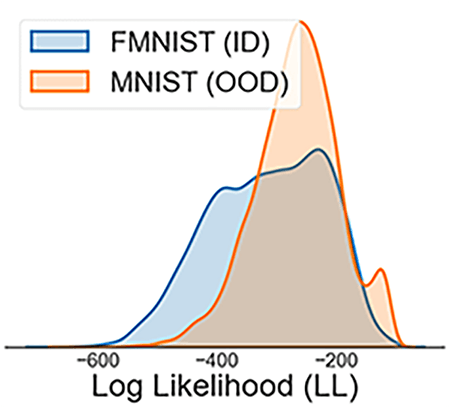}
   \vspace*{-5mm}
   \caption{}

\end{subfigure}
\hspace{2mm}
\begin{subfigure}[b]{0.15\linewidth}
  \includegraphics[width=\textwidth]{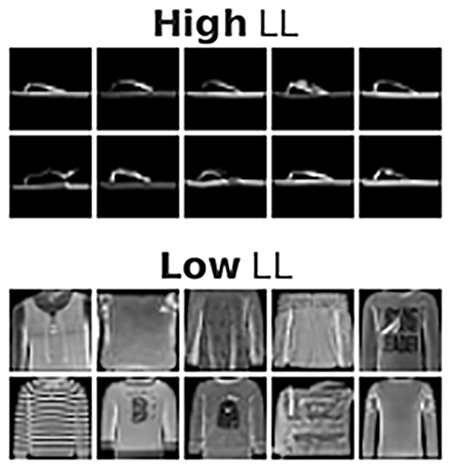}
  \vspace*{-5mm}
  \caption{}

\end{subfigure}
\hspace{2mm}
\begin{subfigure}[b]{0.2\linewidth}
   \includegraphics[width=\textwidth]{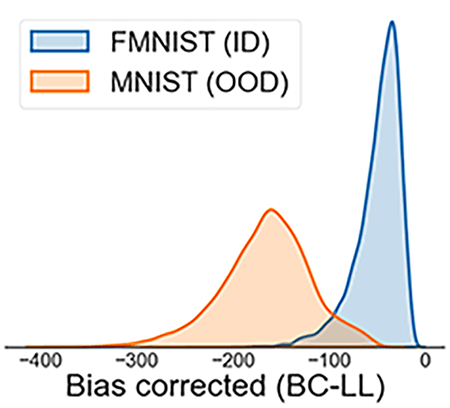}
   \vspace*{-5mm}
   \caption{}

\end{subfigure}
\hspace{2mm}
\begin{subfigure}[b]{0.15\linewidth}
  \includegraphics[width=\textwidth]{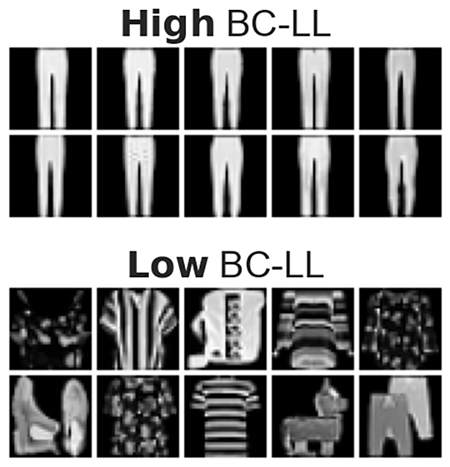}
  \vspace*{-5mm}
  \caption{}

\end{subfigure}
\hspace{2mm}
\begin{subfigure}[b]{0.2\linewidth}
  \includegraphics[width=\textwidth]{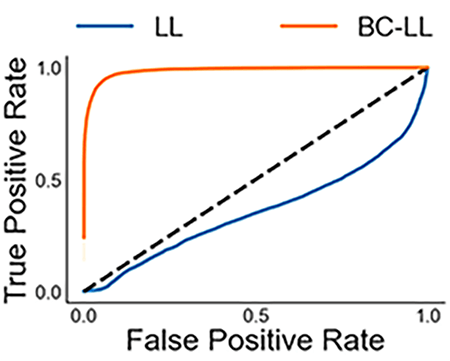}
  \vspace*{-5mm}
  \caption{}

\end{subfigure}

\begin{subfigure}[b]{0.2\linewidth}
   \includegraphics[width=\textwidth]{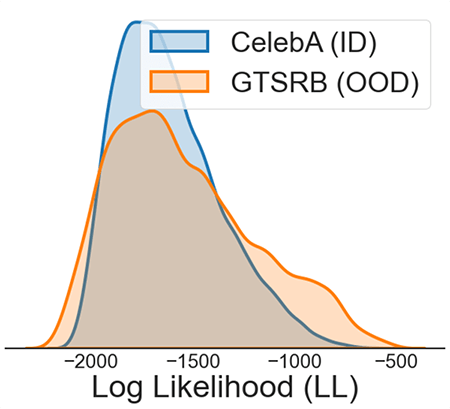}
   \vspace*{-5mm}
   \caption{}

\end{subfigure}
\hspace{2mm}
\begin{subfigure}[b]{0.15\linewidth}
  \includegraphics[width=\textwidth]{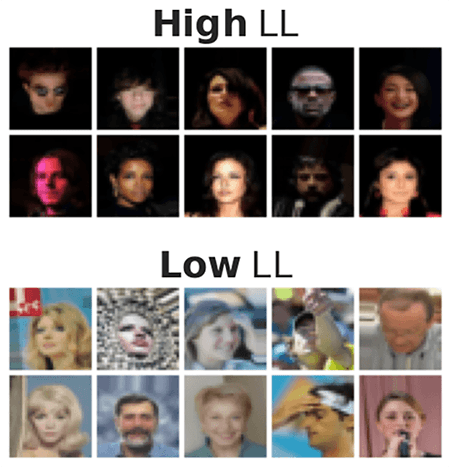}
  \vspace*{-5mm}
  \caption{}

\end{subfigure}
\hspace{2mm}
\begin{subfigure}[b]{0.2\linewidth}
   \includegraphics[width=\textwidth]{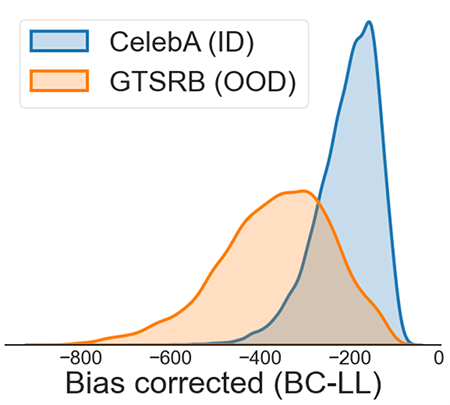}
   \vspace*{-5mm}
   \caption{}

\end{subfigure}
\hspace{2mm}
\begin{subfigure}[b]{0.15\linewidth}
  \includegraphics[width=\textwidth]{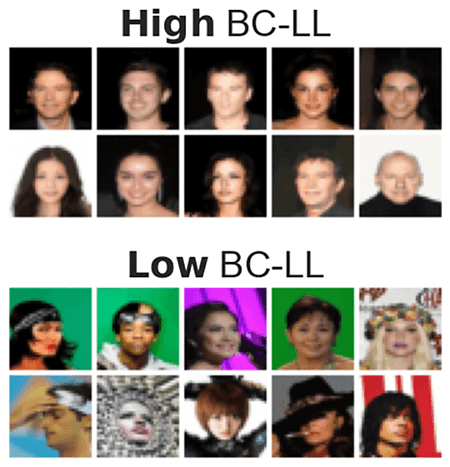}
  \vspace*{-5mm}
  \caption{}

\end{subfigure}
\hspace{2mm}
\begin{subfigure}[b]{0.2\linewidth}
  \includegraphics[width=\textwidth]{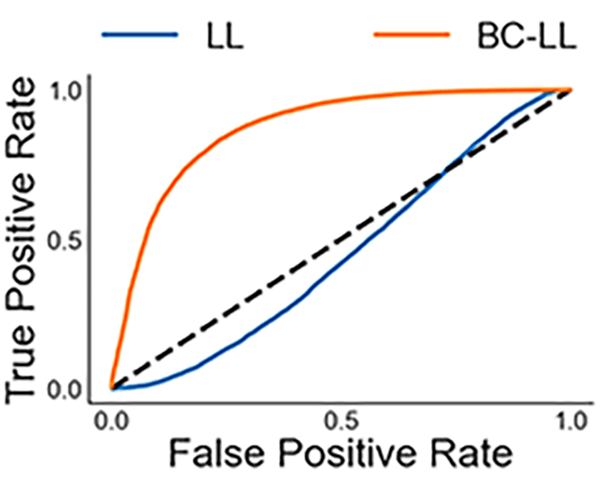}
  \vspace*{-5mm}
  \caption{}

\end{subfigure}
\caption{{\bf Bias correction improves outlier detection for the Bernoulli VAE.} Conventions are same as in Figure~\ref{fig:beforeandafter}, but for a VAE trained with the Bernoulli visible distribution.} 
\label{fig:bern}

\end{figure*}

\begin{figure*}[ht!]
\centering
\begin{subfigure}[b]{0.2\linewidth}
   \includegraphics[width=\textwidth]{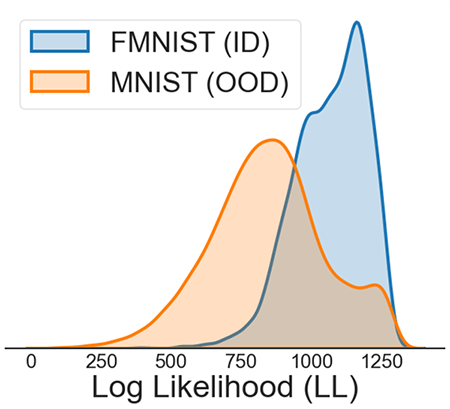}
   \vspace*{-5mm}
   \caption{}

\end{subfigure}
\hspace{2mm}
\begin{subfigure}[b]{0.15\linewidth}
  \includegraphics[width=\textwidth]{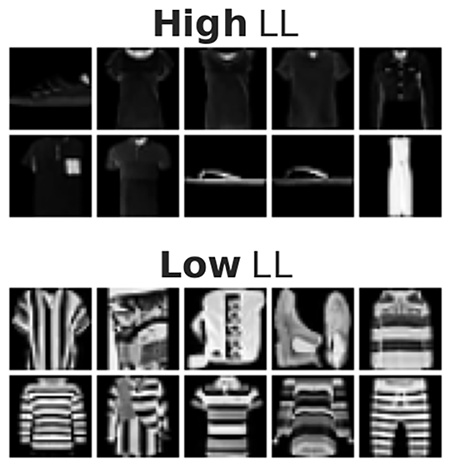}
  \vspace*{-5mm}
  \caption{}

\end{subfigure}
\hspace{2mm}
\begin{subfigure}[b]{0.2\linewidth}
   \includegraphics[width=\textwidth]{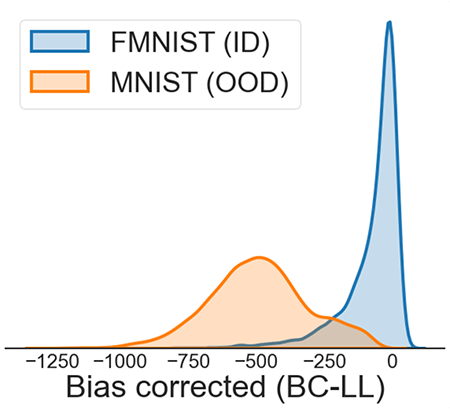}
   \vspace*{-5mm}
   \caption{}

\end{subfigure}
\hspace{2mm}
\begin{subfigure}[b]{0.15\linewidth}
  \includegraphics[width=\textwidth]{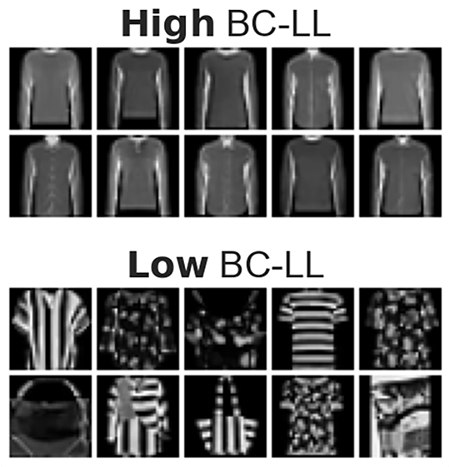}
  \vspace*{-5mm}
  \caption{}

\end{subfigure}
\hspace{2mm}
\begin{subfigure}[b]{0.2\linewidth}
  \includegraphics[width=\textwidth]{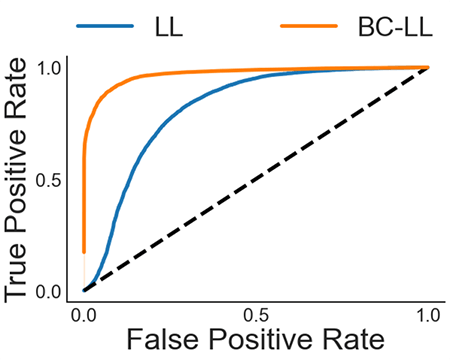}
  \vspace*{-5mm}
  \caption{}

\end{subfigure}

\begin{subfigure}[b]{0.2\linewidth}
   \includegraphics[width=\textwidth]{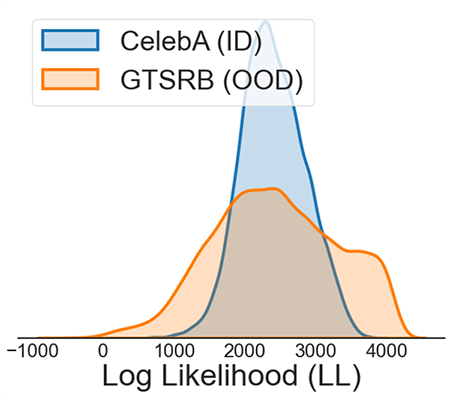}
   \vspace*{-5mm}
   \caption{}

\end{subfigure}
\hspace{2mm}
\begin{subfigure}[b]{0.15\linewidth}
  \includegraphics[width=\textwidth]{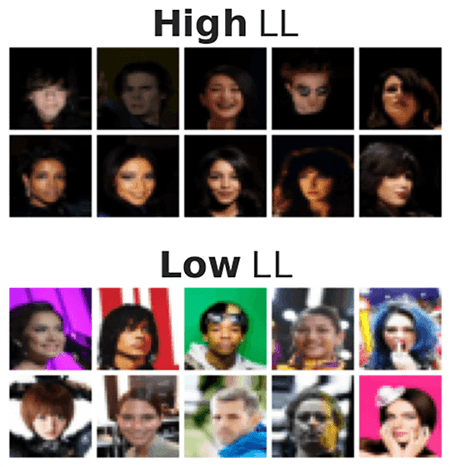}
  \vspace*{-5mm}
  \caption{}

\end{subfigure}
\hspace{2mm}
\begin{subfigure}[b]{0.2\linewidth}
   \includegraphics[width=\textwidth]{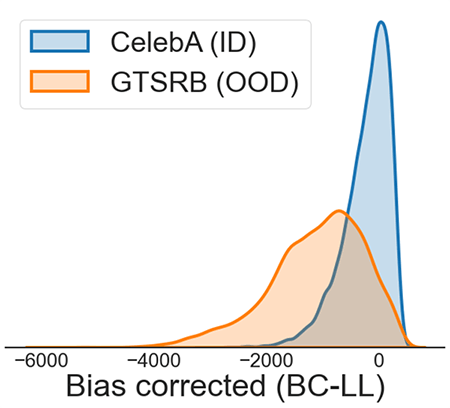}
   \vspace*{-5mm}
   \caption{}

\end{subfigure}
\hspace{2mm}
\begin{subfigure}[b]{0.15\linewidth}
  \includegraphics[width=\textwidth]{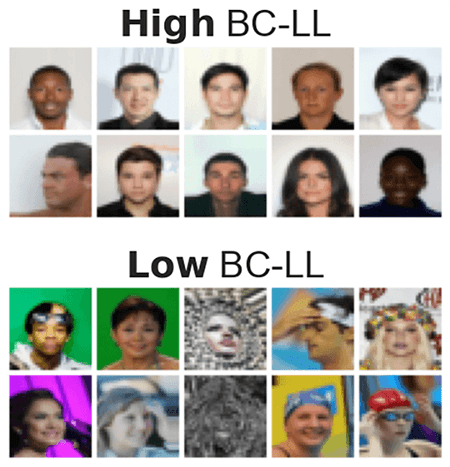}
  \vspace*{-5mm}
  \caption{}

\end{subfigure}
\hspace{2mm}
\begin{subfigure}[b]{0.2\linewidth}
  \includegraphics[width=\textwidth]{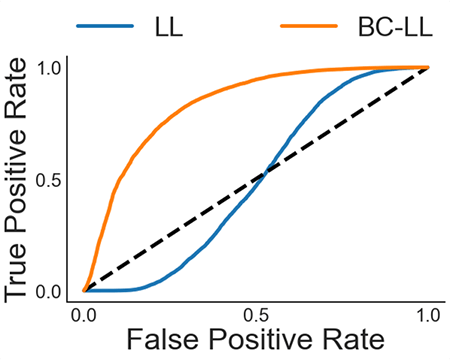}
  \vspace*{-5mm}
  \caption{}

\end{subfigure}
\caption{{\bf Bias correction improves outlier detection for the truncated Gaussian VAE.} Conventions are same as in Figure~\ref{fig:beforeandafter}, but for a VAE trained with the truncated Gaussian visible distribution.} 
\label{fig:gaussian}

\vspace{-1em}
\end{figure*}

\subsection*{F.2 Bias correction  with alternative visible distributions}

We report outlier detection results for Bernoulli (Fig.~\ref{fig:bern}) and truncated Gaussian (Fig.~\ref{fig:gaussian}) VAEs. To correct for the intensity bias in the Bernoulli VAE, we use the analytical approach discussed in Section 3.1 and Appendix B. For truncated Gaussian VAEs, we used the algorithmic correction discussed in Section 3.2 and Algorithm 1. For all VAEs, the images were contrast stretched during both training and testing phases.

\section*{Appendix G: Outlier detection with the CIFAR-10 VAE}

Our BC-LL scores approached or exceeded state-of-the-art accuracies for outlier detection with multiple grayscale and natural image datasets (Figs.~\ref{fig:grayscale} and \ref{fig:natural}). Nonetheless, VAEs trained on the CIFAR-10 dataset yielded relatively low AUROC values with bias correction, ranging from 37-66 (Fig.~\ref{fig:natural}, last column, all outlier datasets except Noise). Interestingly, other approaches based on VAE likelihoods also performed relatively poorly with this dataset (Fig.~\ref{fig:natural}, last column)

\begin{figure}[h]
\centering
   \includegraphics[width=0.85\linewidth]{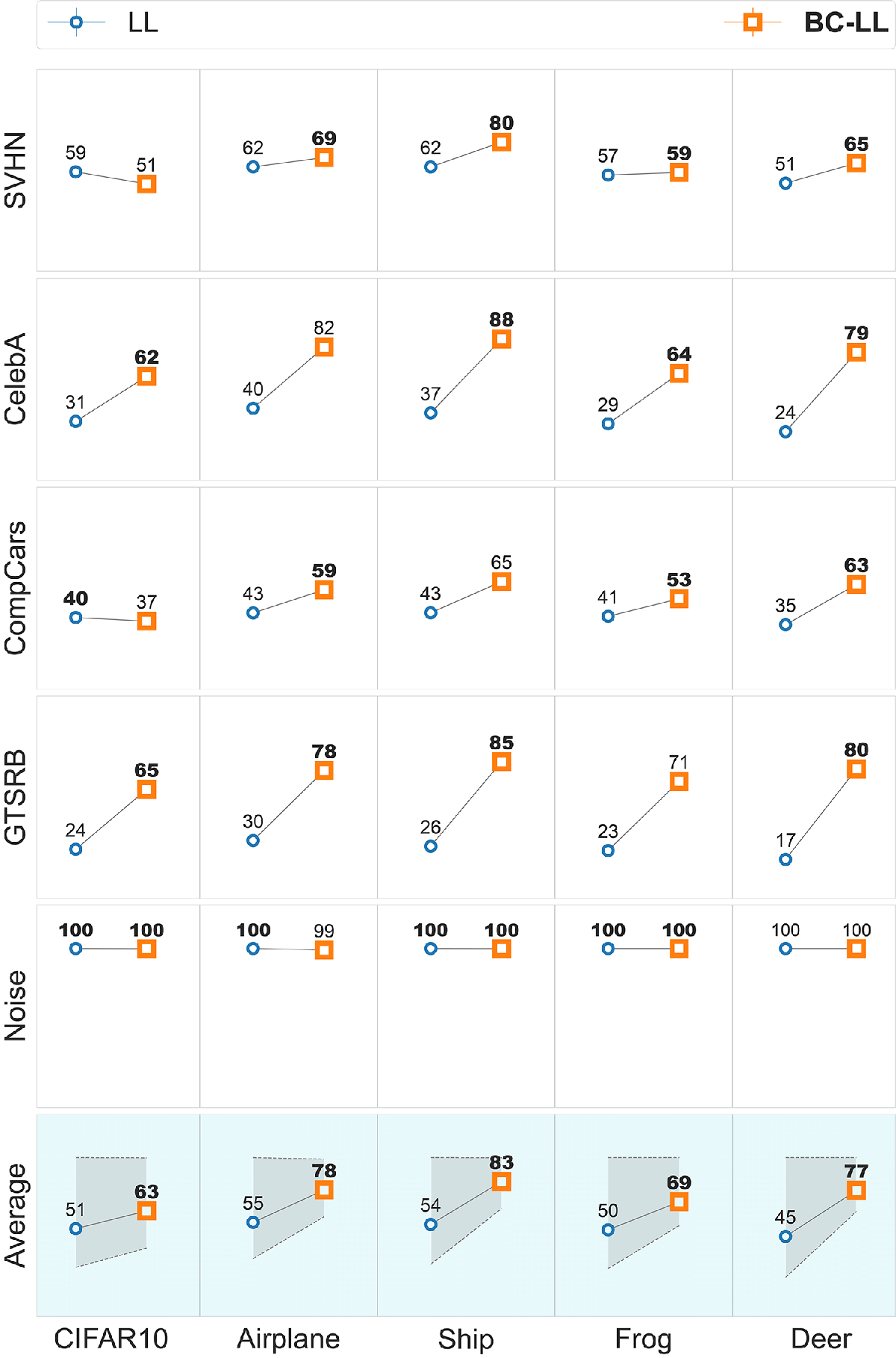}
\caption{{\bf Outlier detection with specific CIFAR-10 categories.} AUROC values for outlier detection based on VAEs trained with specific categories of images in the CIFAR-10 dataset. (Leftmost to rightmost columns) VAEs trained with all CIFAR-10 image categories, Airplane images, Ship images, Frog images and Deer images. Each row is an outlier dataset. Other conventions are the same as in Figure \ref{fig:natural}.} 
\label{fig:cifar10_subcat}
\vspace{-1em}
\end{figure}

We hypothesized that this failure could be due to the heterogeneity of the CIFAR-10 dataset: this dataset comprises 10 different categories of images, which are both visually and semantically unrelated to each other (e.g. airplanes, deer, frogs, ships). We tested this hypothesis by training four category-specific VAEs with images from four CIFAR-10 image categories (Airplane, Ship, Frog and Deer) and tested them against the other datasets. Outlier detection performance improved with bias correction for the VAEs trained on specific categories of images (Fig.~\ref{fig:cifar10_subcat}, last 4 columns) relative to the one trained with all categories of images (Fig.~\ref{fig:cifar10_subcat}, first column). On average, category-specific outlier detection with the BC-LL score improved by between 6-20 points (Fig.~\ref{fig:cifar10_subcat}, last row) for the individual categories compared to the VAE trained on all categories.

These results suggest that the heterogeneity in CIFAR-10 categories was, in part, responsible for overall poor outlier detection with this dataset. Yet, even with these category-specific CIFAR-10 VAEs, outlier detection performance did not reach the superlative levels of accuracy that it did with the other natural image datasets (e.g. Fig.
~\ref{fig:natural}, CompCars, GTSRB).

\begin{figure}[h]
\centering
   \includegraphics[width=1\linewidth]{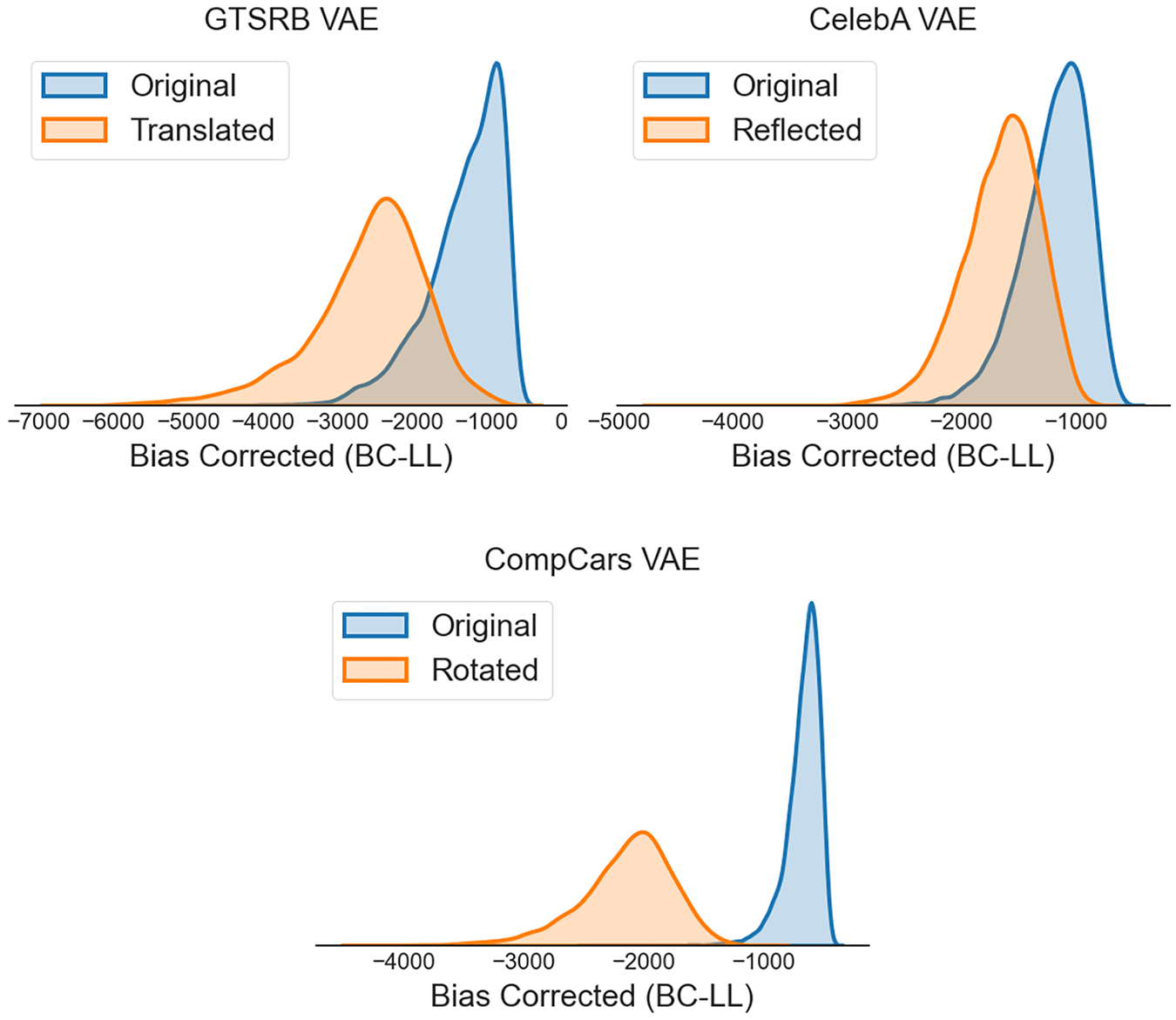}
\caption{{\bf VAE likelihoods following affine transformations to input data}.
{\bf (Top Left)} GTSRB VAE likelihoods (bias corrected) for the original images (blue distribution) or for the images following small, random wrap-around x/y translations (orange distribution).
{\bf (Top Right)} CelebA VAE likelihoods (bias corrected) for the original images (blue distribution) or for the images flipped vertically (orange distribution). 
{\bf (Bottom)} CompCars VAE likelihoods (bias corrected) for the original images (blue distribution) or for the images rotated by 90 degrees anti-clockwise (orange distribution).
} 
\label{fig:bcll_flip}
\vspace{-1em}
\end{figure}


{\it Why do VAE likelihoods fail when other classifier-based outlier detection methods (e.g.\cite{Lakshminarayanan2017}) succeed?} To answer this question, we note that there is a key difference between VAEs and deep CNN classifiers. Deep CNN classifiers transform the input image into a spatially-invariant (typically translation-invariant) feature representation, on which a classification decision must be made. Even though the VAE encoder employs a CNN that achieves such a transformation, the VAE decoder needs to reconstruct the entire image pixel-for-pixel. Consequently, VAEs can ill-afford to ignore spatial relationships among the features and their positions relative to a coordinate frame that is locked to the ``edge'' of the image. In other words, to optimize its objective, a VAE must care not only about ``what'' features are present in an image, it must also encode ``where'' these features are positioned relative to each other and, importantly, relative to the image's bounding box. We propose that datasets in which such spatial relationships do not occur consistently (e.g. CIFAR-10) are particularly unsuited for outlier detection with VAEs.

We test this prediction with a simple set of experiments. We plotted the bias corrected likelihoods for representative VAEs (GTSRB, CelebA, CompCars), each following one simple affine transformation: a) x/y translation (uniform random shift of $\pm$0-10 pixels along both x and y directions) (Fig.~\ref{fig:bcll_flip}, top left, GTSRB), b) reflection about the horizontal axis (Fig.~\ref{fig:bcll_flip}, top right, CelebA) or c) rotation by 90 degrees anti-clockwise (Fig.~\ref{fig:bcll_flip}, bottom, CompCars). Each of these affine operations sufficed to hoodwink the VAEs into treating their, respective, inlier images as outliers: the VAEs consistently assigned lower likelihoods for these translated, reflected or rotated in-distribution images (Fig.~\ref{fig:bcll_flip}, orange distributions), as compared to their original counterparts (Fig.~\ref{fig:bcll_flip}, blue distributions). This was particularly surprising for the case of perturbation by translation, to which the encoder CNN typically learns invariance. In sum, even though all image features were exactly (or nearly) identical between the original and affine transformed images, VAE outlier detection failed because perturbed image features were in novel spatial positions relative to the edge of the image.


We discuss a few possibilities to overcome this limitation. One solution is to train the VAE, not directly on natural images, but after transforming them into a spatially-invariant ``semantic'' feature representation. Indeed, a recent study \cite{Krichenko2020} attempted precisely this experiment with deep generative (Flow) models using features from a pretrained EfficientNet model, and showed that such features enabled robust outlier detection, even with the CIFAR-10 dataset. A second possibility is to perturb the output \cite{Hendrycks2019} and develop reconstruction error metrics that are invariant to these perturbations in the input data. These remedies could enable VAEs to be employed in dynamic real-world settings where objects of interest may not be stationary relative to the edge of the bounding frame, for example, outlier detection in video data.
\end{document}